%% file: paper_template.tex
\documentclass[conference]{IEEEtran}
\usepackage{times}

\usepackage[numbers]{natbib}
\usepackage{multicol}
\usepackage[bookmarks=true]{hyperref}
\usepackage[table]{xcolor} 
\usepackage{graphicx}
\usepackage{url}
\usepackage{subcaption}
\usepackage{pdfpages}
\usepackage{booktabs}
\usepackage{balance}
\usepackage{tabularx}
\usepackage{adjustbox}
\usepackage{comment}
\usepackage{fancyhdr} 
\usepackage{xspace} 
\usepackage{wrapfig}
\usepackage[many]{tcolorbox}
\usepackage{amsmath}
\usepackage{amsfonts} 
\usepackage{cleveref}
\usepackage{algorithm}
\usepackage[noend]{algpseudocode}
\usepackage[numbers]{natbib}
\usepackage{listings}
\usepackage{pifont}
\usepackage{sidecap}

\let\labelindent\relax 
\usepackage{enumitem}

\definecolor{codebackground}{rgb}{1.0, 1.0, 1.0} 
\definecolor{codeblue}{rgb}{0.1, 0.5, 0.9}          
\definecolor{codegreen}{rgb}{0.1, 0.5, 0.1}         
\definecolor{codegray}{rgb}{0.5, 0.5, 0.5}          
\definecolor{codered}{rgb}{0.8, 0.1, 0.1}           
\definecolor{advisorgreen}{RGB}{136,189,117}
\definecolor{groundingyellow}{RGB}{255,224,133}
\definecolor{roboticblue}{RGB}{123,173,244}
\definecolor{monitorgray}{RGB}{183,183,183}

\newcommand{\xxnote}[3]{}
\ifx\hidenotes\undefined
\renewcommand{\xxnote}[3]{\color{#2}{#1: #3}}
\fi

\definecolor{codegreen}{rgb}{0,0.6,0}
\newcommand{\CommentColored}[1]{\Comment{\textcolor{codegreen}{\textit{#1}}}}
\newcommand{\CommentColoredR}[1]{
    \begin{flushright}
    \textcolor{codegreen}{\textit{#1}}
    \end{flushright}
}

\newcommand{\cut}[1]{}
\newcommand{\para}[1]{{\noindent\textbf{#1}}}
\newcommand{\parab}[1]{{\noindent\textbf{#1}}}
\newcommand{\parai}[1]{{\noindent\textit{#1}}}

\renewcommand{\cite}{\citep}

\newcommand{\ourmethod}{\texttt{GRAPPA}\xspace}

\newcommand{\cmark}{{\color{green}\ding{51}}}  
\newcommand{\xmark}{{\color{red}\ding{55}}}  

\newtcolorbox{llmprompt}[3][]{%
    enhanced jigsaw, 
    breakable,      
    left=1cm,       
    right=1cm,      
    colframe    = #2!25,
    colback     = #2!10,
    coltitle    = #2!20!black,  
    title       = {#3},
    parbox=false,
}

\hypersetup{
    colorlinks=true,
    citecolor=[RGB]{0,0,0},
    linkcolor=[RGB]{0,0,0},
    filecolor=[RGB]{0,0,0},      
    urlcolor=[RGB]{0,0,0},
    pdfpagemode=FullScreen,
}

\begin{document}

\title{\textit{\ourmethod}: Generalizing and Adapting Robot Policies via Online Agentic Guidance}


\author{Arthur Bucker$^{1}$, Pablo Ortega-Kral$^{1}$, Jonathan Francis$^{1,2}$, Jean Oh$^{1}$ \\  
$^{1}$Robotics Institute, Carnegie Mellon University \\
$^{2}$Robot Learning Lab, Bosch Center for Artificial Intelligence\\ 
{\tt\footnotesize \{abucker, portegak, jmf1, jeanoh\}@cs.cmu.edu}
}



%

\makeatletter
\let\@oldmaketitle\@maketitle
\renewcommand{\@maketitle}{\@oldmaketitle
\centering
  \includegraphics[width=0.9\linewidth]{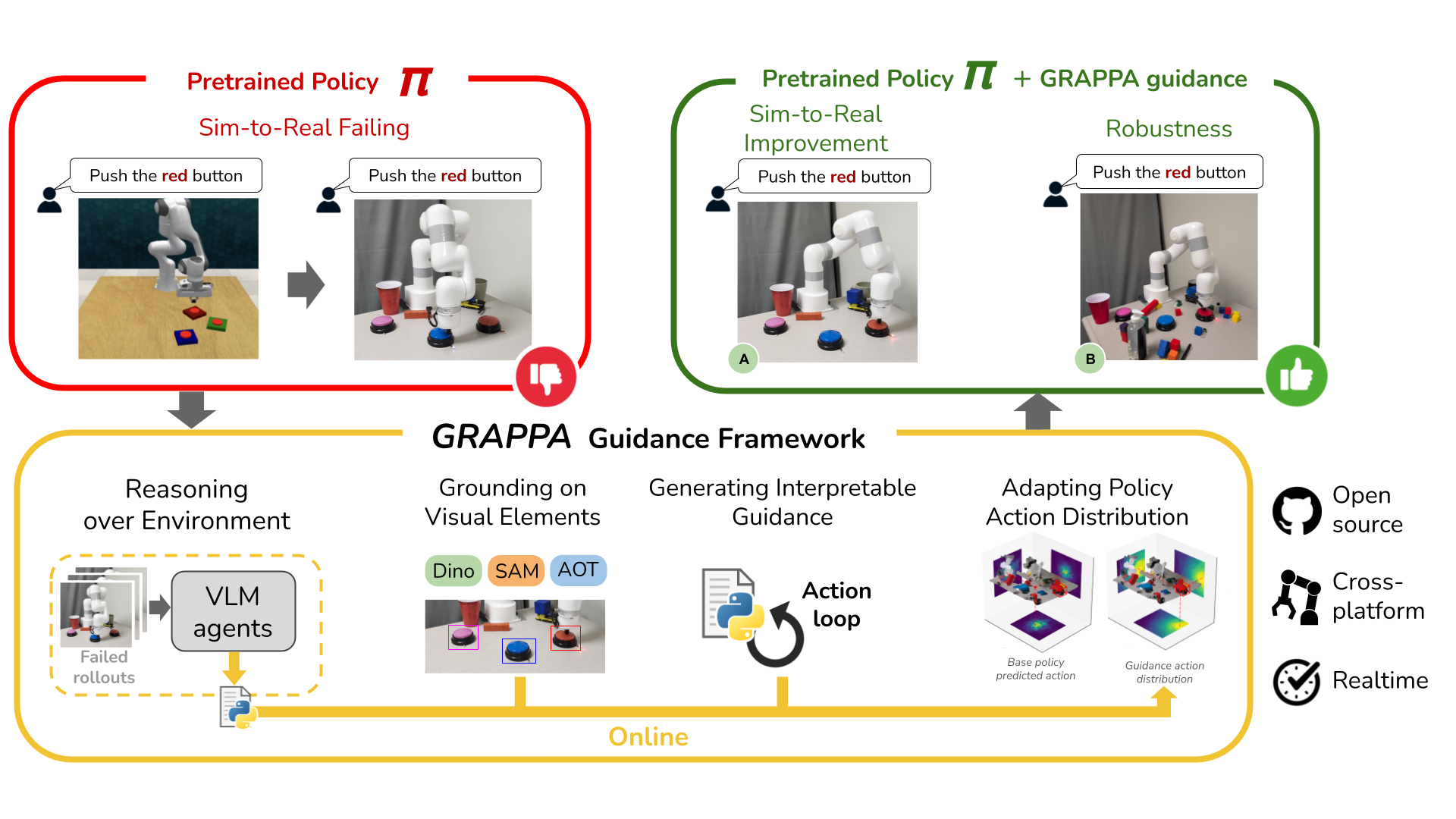}
  \captionof{figure}{\textbf{\ourmethod framework.} An overview of the proposed self-improving framework where the pre-trained policy is updated by visuomotor-grounded guidance at test time. \ourmethod can guide and provide robustness to policies under two main failure modes: A) when a pre-trained policy with limited to no access to real data is applied to an embodied system and B) there is a lot of noise, in the form of clutter, affecting the observations of the base policy.} 
  \label{fig:splash}
  \vspace{-0.07in}
}
\makeatother

\let\oldmaketitle\maketitle
\renewcommand{\maketitle}{
  \oldmaketitle
  \addtocounter{figure}{-1}  
}

\maketitle

\input{inputs/0_abstract}

\IEEEpeerreviewmaketitle

\input{inputs/1_introduction}
\input{inputs/2_related_works}
\input{inputs/3_methodology}
\input{inputs/4_experiments_results}

\input{inputs/5_discussion}
\bibliographystyle{plainnat}
\bibliography{references}
\clearpage
\input{inputs/6_appendix}

\end{document}

%% file: inputs/0_abstract.tex


\begin{abstract}
Robot learning approaches such as behavior cloning and reinforcement learning have shown great promise in synthesizing robot skills from human demonstrations in specific environments. However, these approaches often require task-specific demonstrations or designing complex simulation environments, which limits the development of generalizable and robust policies for unseen real-world settings.
Recent advances in the use of foundation models for robotics (e.g., LLMs, VLMs) have shown great potential in enabling systems to understand the semantics in the world from large-scale internet data. However, it remains an open challenge to use this knowledge to enable robotic systems to understand the underlying dynamics of the world, to generalize policies across different tasks, and to adapt policies to new environments.
To alleviate these limitations, we propose an agentic framework for robot self-guidance and self-improvement, which consists of a set of role-specialized conversational agents, such as a high-level advisor, a grounding agent, a monitoring agent, and a robotic agent. Our framework iteratively grounds a base robot policy to relevant objects in the environment and uses visuomotor cues to shift the action distribution of the policy to more desirable states, online, while remaining agnostic to the subjective configuration of a given robot hardware platform. 
We demonstrate that our approach can effectively guide manipulation policies to achieve significantly higher success rates, both in simulation and in real-world experiments, without the need for additional human demonstrations or extensive exploration. Code and videos available at: \url{https://agenticrobots.github.io}

\end{abstract}

%% file: inputs/1_introduction.tex
\section{Introduction}

In recent years, the advent of foundation models, such as pre-trained large language models (LLMs) and vision language models (VLMs), has enabled impressive capabilities in understanding context, scenes, and dynamics in the world. Furthermore, emergent capabilities such as in-context learning have shown great potential in the transfer of knowledge between domains, e.g., via few-shot demonstrations or zero-shot inference. However, the application of these models to robotics is still limited, given the intrinsic complexity and scarcity of human-robot interactions and the lack of large-scale datasets of human-annotated data or demonstrations \cite{hu2023toward, yenamandra2024towards}.

Approaches that leverage LLMs and VLMs alongside robotics systems often fall into one of two categories. The first category is that of using foundation models as zero-shot planners, code-generators, and task-descriptors---all of which attempt to use the foundation model to provide some high-level instruction to a low-level policy or to describe the policy's actions in natural language \cite{rana2023sayplan, saxena2024grapheqa, ahn2022can,liu2023reflect} or generated plans as code \cite{liang2023code, singh2023progprompt}. While these works have illustrated impressive reasoning capabilities, they still required either learning additional mappings to interface the generated natural language instructions with the low-level policy or, in the case of the code generation examples, they rely on pre-existing handcrafted primitives to compose. For the second category, in the use of foundation models for robot learning, methods may fine-tune the foundation models to improve in-domain performance by imparting strong priors for effective, albeit platform-dependent task-execution \citep{brohan2023rt2, team2024octo, kim2024openvla}. Here, parts of a robot policy or the entire policy is supervised to perform certain prediction tasks or to learn desired behaviors, end-to-end. While this strategy has led to impressive capabilities on a variety of tasks, methods often struggle to generalize to different object categories, tasks, and environments that are outside the data distribution that the models were trained on, or the models struggle with issues of negative transfer when trained on very diverse collections of robot data \citep{open_x_embodiment_rt_x_2023, khazatsky2024droid}.

In this paper, we extend the deployability of robot policies by using foundation models to generate low-level visuomotor guidance to handle out-of-distribution scenarios, such as sim-to-real differences or new tasks and robot platform variations (Figure \ref{fig:close_up_guidance}). We design an agent-based framework, where a team of conversational agents works together to refine the action distribution of a robot's base policy, via grounded visuomotor guidance (Figure~\ref{fig:system_overview}). As illustrated in Figure~\ref{fig:information_flow}, and upon a request from the advisor agent, the grounding agent can look for objects through a combination of detection, tracking, and high-level reasoning, to broaden or restrict the search as the context demands. 
When looking for a mug, for example, the agents could first look for a semantically-relevant receptacle that likely contains the mug (the cupboard) and narrow the search from there. Once the target object has been visually located, the Monitor and Advisor agents generate a guidance function that outputs a biasing guidance distribution, which, when combined with the action distribution of the Robotic agent (policy), ensures that the policy can complete task-relevant behaviors until it succeeds. In this way, the agentic framework bridges high-level reasoning with low-level control, enabling systems to reason about failure and self-guide. 



In simulated and real-world experiments, we comprehensively and empirically demonstrate that our framework for \textbf{G}ene\textbf{r}alizing and \textbf{A}da\textbf{p}ting Robot \textbf{P}olicies via Online \textbf{A}gentic Guidance (\ourmethod) provides robustness for a variety of representative base policy classes to sim-to-real transfer paradigms and out-of-distribution settings, such as unseen objects. 


In summary, our paper provides the following contributions:
\begin{itemize}[noitemsep, left=0pt, topsep=0pt]
    \item We propose \ourmethod, an agentic robot policy framework that self-improves by generating a guidance function to update base policies' action distributions, online. Our framework is capable of learning skills from scratch after deployment and generalizes across various base policy classes, tasks, and environments.
    \item For robustness against cluttered and unseen environments, we propose a grounding agent that performs multi-granular object search, which enables flexible visuomotor grounding. 
    \item We provide experimental results showcasing \ourmethod as a helper tool for aiding in the sim-to-real transfer of policies without the need for extensive data collection.
\end{itemize}


%% file: inputs/2_related_works.tex
\begin{figure}[!t]
    \centering
    \label{fig:heatmap_sim_to_real}
    \includegraphics[width=\columnwidth]{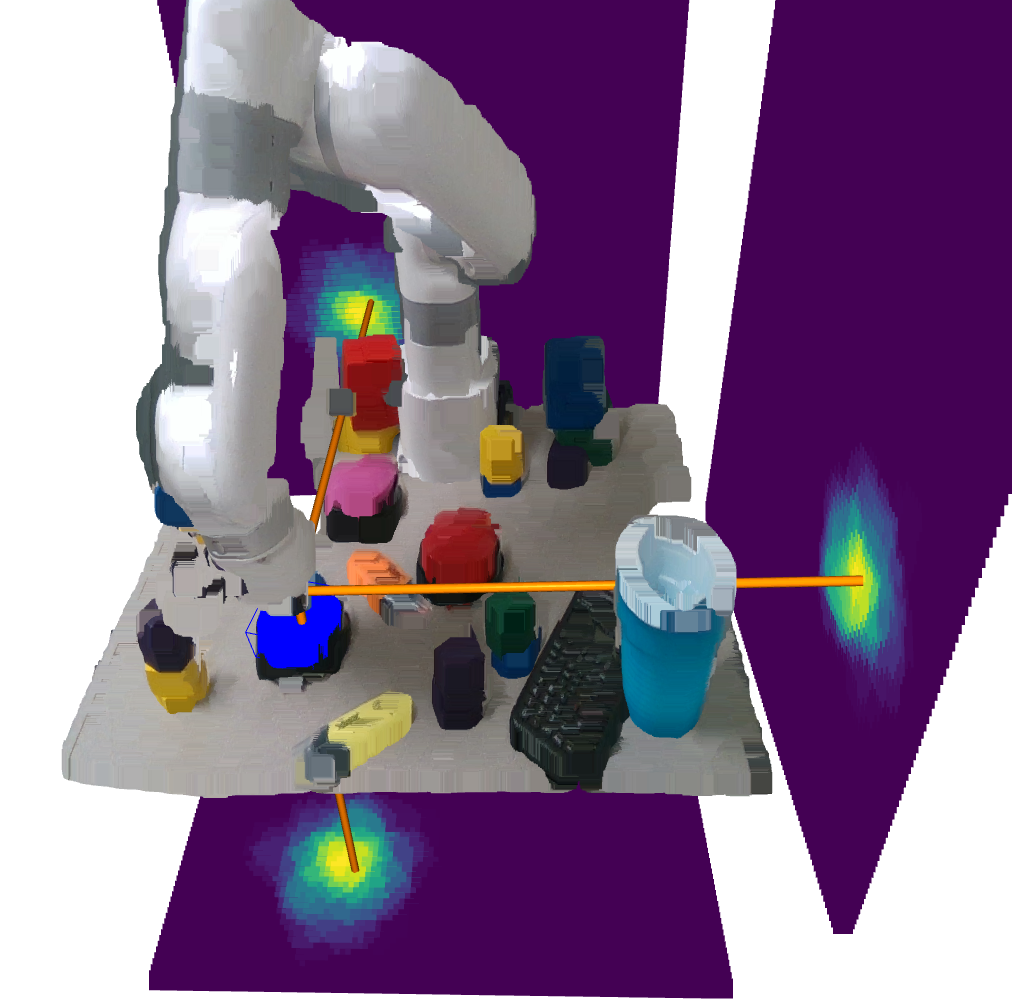}
    \caption{Heatmap visualization of the guidance distribution, generated online by our proposed agentic framework, which produces code that biases a robot policy's action distribution towards desirable behavior.} \label{fig:close_up_guidance}
    \vspace{-0.5cm}
\end{figure}

\begin{figure*}[h]
    \centering
    \includegraphics[width=0.9\textwidth]{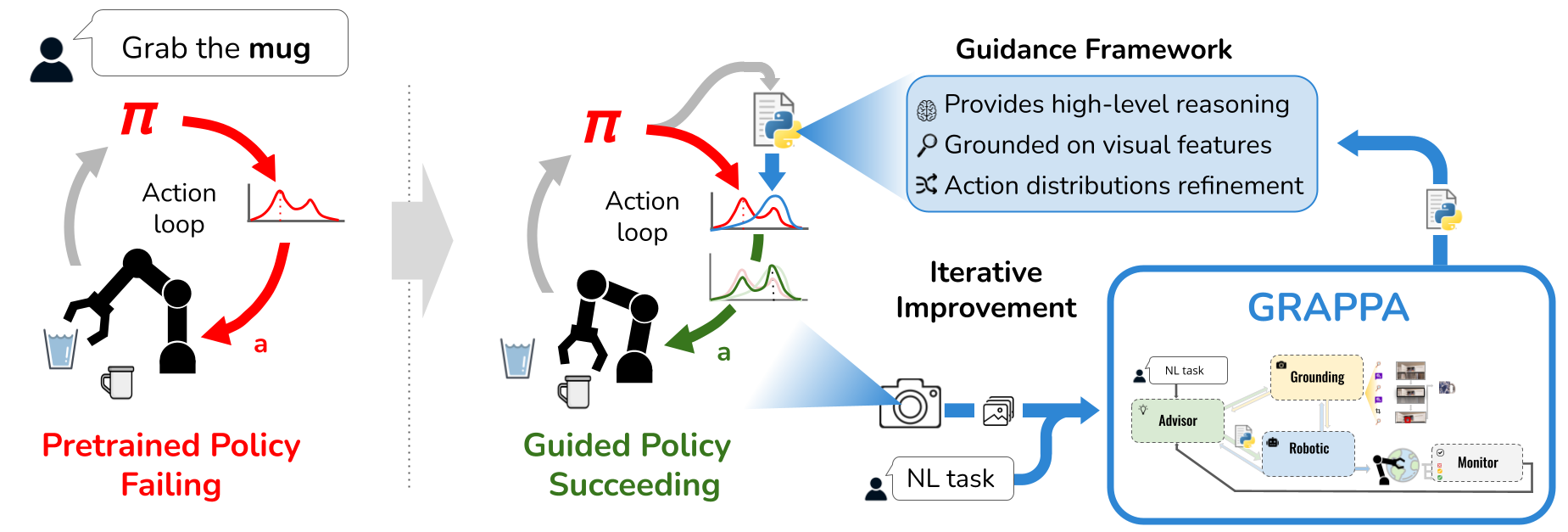}

    \caption{An illustration of how \ourmethod intervenes in the action loop of pre-trained robotic policies in failure cases to provide visuomotor guidance generated with an agentic framework of agents to shift the action distribution for correct task execution. }%
    \label{fig:system_overview}
\end{figure*}

\section{Related Work}
\label{sec:related_work}

\parab{Vision-Language Models for Robot Learning:} Several works explore the notion of leveraging pre-trained or fine-tuned Large Language Models (LLMs) and/or Vision-Language Models (VLMs) for high-level reasoning and planning in robotics tasks \citep{hu2023toward, ahn2022can, liang2023code, huang2022inner, singh2023progprompt, huang2023visual, ha2023scaling, ding2023task, michal2024robotic, ma2023eureka, li2024manipllm, mu2024embodiedgpt}---typically decomposing high-level task specification into a series of smaller steps or action primitives, using system prompts or in-context examples to enable powerful chain-of-thought reasoning techniques. This strategy of encouraging models to reason in a stepwise manner before outputting a final answer has led to significant performance improvements across several tasks and benchmarks \citep{hu2023toward}. Despite these promising achievements, these approaches rely on handcrafted primitives \citep{ahn2022can, huang2022inner, liang2023code, michal2024robotic}, struggle with low-level control, or require large datasets for retraining. 
 Furthermore, various approaches that leverage VLMs for robot learning suffer from a granularity problem when using off-the-shelf models in a single-step/zero-shot manner \citep{huang2023voxposer} or are unable to perform failure correction without costly human intervention \citep{huang2022inner, michal2024robotic, liang2024learning, huang2023voxposer}. In contrast, our framework bridges high-level reasoning with low-level control, by leveraging an agentic framework for online modification of a base policy's action distribution at test-time, without requiring human feedback or datasets for fine-tuning. Moreover, we mitigate the granularity problem by proposing a flexible and recursive grounding mechanism that uses VLMs to query open-vocabulary perception models.


\parab{Agent-based VLM frameworks in Robotics:} Rather than using single VLMs in an end-to-end fashion, which might incur issues in generalization and robustness, various works have sought to orchestrate multiple VLM-based agents to work together in an interconnection multi-agent framework. Here, multiple agents can converse and collaborate to perform tasks, yielding improvements for the overall framework in online adaptability, cross-task generalization, and self-supervision \citep{xu2023creative, zhang2024rail, parakh2023lifelong}. These \textit{agentic} frameworks have provided possibilities for enabling the identification of issues in task execution, providing feedback about possible improvements. Challenges remain, however, in that this feedback is often not sufficiently grounded on the spatial, visual, and dynamical properties of embodied interaction to be useful for policy adaptation; instead, the generated feedback is often too high-level or provides merely binary signals of success or failure. 

\parab{Self-guided Robot Failure Recovery:} \citet{guan2024task} offer an analysis of frameworks for leveraging VLMs as behavior critics. Some approaches have explored integrating such pre-trained models to improve the performance of reinforcement learning (RL) algorithms. For instance, \citet{ma2023eureka} use LLMs in a zero-shot fashion to design and improve reward functions, however this approach relies on human feedback to generate progressively human-aligned reward functions and further requires simulated retraining via RL, with high sample-complexity. On the other hand, \citet{rocamonde2023vision} avoid the need for explicit human feedback by directly using a VLM (CLIP) to compute the rewards to measure the proximity of a state (image) to a goal (text description), enabling gains in sample-efficiency for guiding a humanoid robot to perform various maneuvers in the MuJoCo simulator. A limitation of this approach, however, lies in the difficulty of generating rewards for long-horizon or multi-step tasks
, which are characteristic of tasks involving complex agent-object interactions.
\citet{liu2023reflect} present a framework for detecting and analyzing failed executions automatically. However, their system focuses on explaining failure causes and proposing suggestions for remediation, as opposed to also performing policy correction. In this paper, we propose a framework that directly adapts a base policy's action distribution, during deployment, without requiring additional human feedback.

%% file: inputs/3_methodology.tex
\begin{figure*}[ht]
    \centering%
    \includegraphics[width=\textwidth]{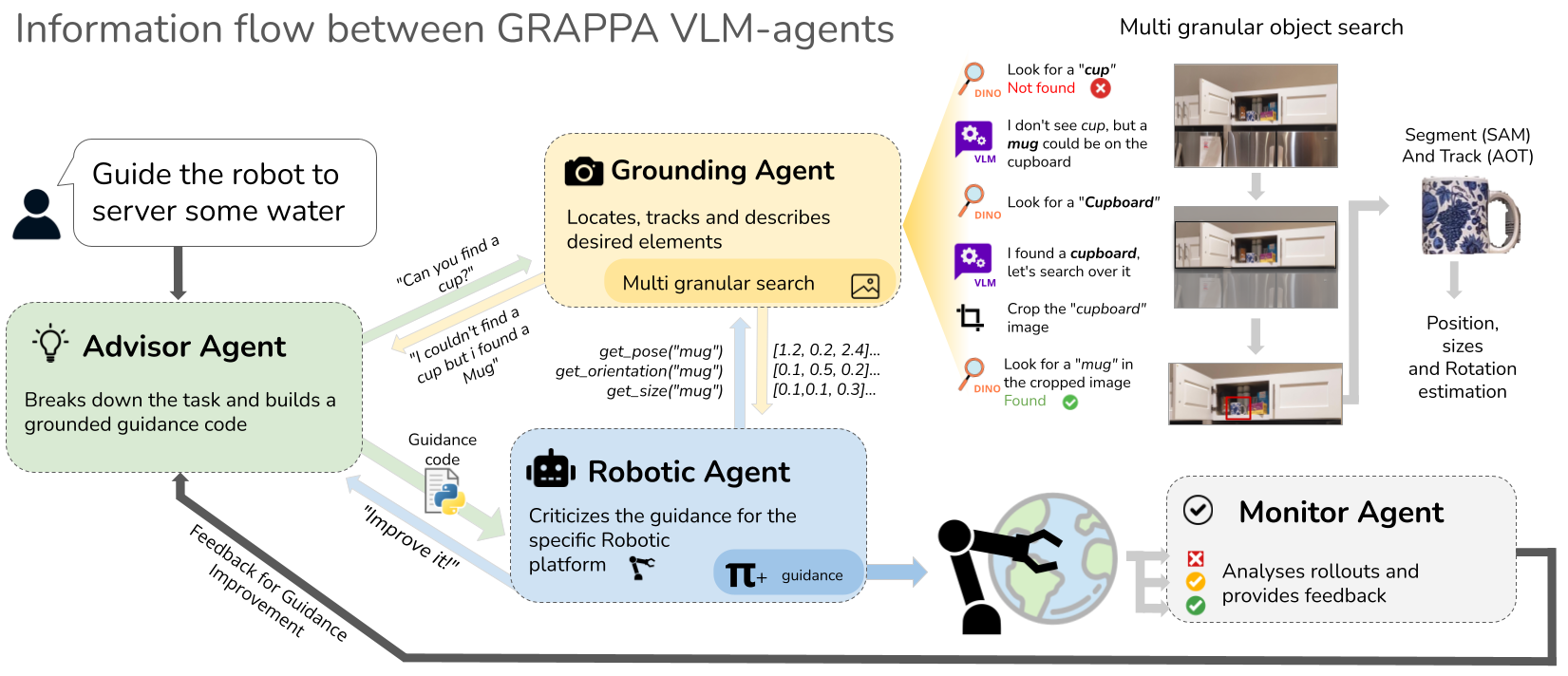}%
    \caption{Information flow between the agents to produce a guidance code.  \textbf{a)} The advisor agent orchestrates guidance code generation by collaborating with other agents and using their feedback to refine the generated code. \textbf{b)} The grounding agent uses segmentation and classification models to locate objects of interest provided by the advisor, reporting findings back to the advisor. \textbf{c)} The robotic agent uses a Python interpreter to test the code for the specific robotic platform and judge the adequacy of the code. \textbf{d)} The monitor agent analyses the sequence of frames corresponding to the rollout of the guidance and give feedback on potential improvements. }%
    \label{fig:information_flow}%
\end{figure*}

\section{Grounding Robot Policies with Guidance}
\label{sec:methodology}

\subsection{Problem Formulation}

We consider a pre-trained stochastic policy $\pi: O \times S \rightarrow A$ that maps observations $o_t$ and robotic states $s_t$ to action distributions $a_{\pi,t}$, at each time step $t$. Our objective is to generate a guidance distribution $g_t$ that, when combined with this base policy, enhances overall performance during inference without requiring additional human demonstrations or extensive exploration procedures. Specifically, we aim to develop a modified policy $\pi_{\text{guided}}: O \times S \rightarrow A$ that achieves better performance on tasks where the original policy $\pi$ struggles. We define this new policy as follows:
\begin{equation}
    \pi_{guided}(a_{g,t} | o_t, s_t) = \pi(a_{\pi,t} | o_t, s_t) * G( a_{\pi,t} | o_t, s_{t+1}'),
\end{equation}
where $G: A \times O \times S \rightarrow [0,1]$ is a guidance function that maps observation $o_t$, action $a_t$, possible future state $s_{t+1}'$ into a guidance score $g_t$. The `$*$' operator here denotes the operation of combining both distributions conceptually, which we explore in detail in Section \ref{subsec:guidance_policy_integration}. 
For the scope of this project, we assume that a dynamics model $\mathcal{D}: S \times A \rightarrow S$ is available, which can forecast possible future states of the robot $s_{t+1}' = \mathcal{D}(s_{t}, a_{\pi,t})$ given the current state $s_t$ and action $a_{\pi,t}$. 

Focusing on leveraging the world knowledge of Vision Language Models, while avoiding adding latency to the action loop, we choose to express these guidance functions as Python code. By integrating these code snippets into action loop of the base policies, we eliminate the need of time-consuming queries to large reasoning models. Samples of the format and content of the guidance functions generated by the framework are presented in the Appendix \ref{subsec:guidance_code_samples}.


\subsection{A Multi-agent Guidance framework for Self Improvement}


In order to generate the guidance function $G$, we leverage a group of conversational agents empowered with visual grounding capabilities and tool usage. Illustrated in Figure \ref{fig:information_flow}, the framework is composed of four main agents: an Advisor Agent, the Grounding Agent, the Monitor Agent, and the Robotic Agent. We provide system prompt samples in Appendix \ref{sec:app:prompts}.

\parai{Advisor Agent.} A Vision Language Model is responsible for breaking down the task and communicating with the other agents to generate a sound guidance function for a given task. 

\parai{Grounding Agent.} A Vision Language Model that iteratively queries the free-form text segmentation models to locate \citep{cheng2023segment, liu2023grounding}, track \citep{cheng2023segmenttrack}, and describe elements relevant to the task execution.




\parai{Monitor Agent.} Responsible for identifying the causes of the failures in the unsuccessful rollouts, the Monitor Agent consists of a Vision Language model equipped with a key frame extractor.

\parai{Robotic Agent.} Language Model equipped with descriptions of the robot platform, a robot's dynamics model and wrapper functions for integration with the base policy. It criticizes the provided guidance functions to reinforce its relevance to the task and alignment with the robot's capabilities. 


\subsection{Guidance Procedure}

The conversational agents interact with each other through natural language and query their underlying tools to iteratively produce a guidance code tailored to the task in hand, the environment, and the robot's capabilities. The information flow between these agents is depicted by Figure \ref{fig:information_flow}. 

For a given task expressed in natural language and an image of the initial state of the environment, the \textit{Advisor Agent} uses Chain-of-Tought \citep{wei2023chainofthoughtpromptingelicitsreasoning} strategy to generate a high-level plan of the steps necessary to accomplish the task. Being able to query a \textit{Grounding Agent} and the \textit{Robotic agent}, the Advisor is able to collect relevant information about trackable objects and elements in the environment, as well as the capabilities and limitations of the robotic platform.

For a given plan and list of relevant objects required for the task completion, the Grounding Agent uses grounding Dino \citep{liu2023grounding} and the Segment Anything Model (SAM) \citep{cheng2023segment} to locate the elements across multiple granularities and levels of abstraction. For instance, if an object is not immediately found, the agent will actively look for semantically similar objects or will look for higher-level elements that could encompass the missing object. For example, if the object ``cup" is to be located, and it could not be immediately found, the agent could search for similar object like a ``mug". If it still struggles to locate it, the agent could search for a ``shelf" and then try to find the ``cup" or ``mug" in the cropped image of the ``shelf". If an object is found, it is added to a tracking system \citep{yang2022deaot}. This process enhances the Segment and Track Anything \citep{cheng2023segment} approach with flexible multi-granular search. The object statuses are reported back to the Advisor Agent, which iterates on the action plan or proceeds with the generation of a guidance function grounded on the trackable objects located.

The Robotic agent acts as a critic to improve the guidance function generated. Equipped with a Python interpreter and details of the base policy's action space, the agent can evaluate the guidance function in terms of feasibility and relevance to the robot's capabilities. Once a function suffices the system's requirements, it is saved to be used in the action loop, in combination with the dynamics model, to provide a guidance score for possible actions sampled from the base policy.

After the execution of a rollout and the identification of failure in the task completion, the Monitor Agent is triggered to analyze the causes of the failure. By extracting key frames from the rollout video using PCA \citep{mackiewicz1993principal} and K-means clustering, the agent can feed a relevant and diverse set of images to the Visual Language Model prompted to access the failure causes. In the iterative applications of our framework, the Monitor Agent provides this feedback to the Advisor Agent, which can use this information to refine the guidance functions generated in the previous iterations.

\para{Temporal-aware Guidance Functions.} Inspired by recurrent architectures, we instruct the agents to generate guidance function conditioned on a customizable hidden state ($h_t$) expressed as an optional dictionary parameter as shown in the following example:

\begin{lstlisting}[language=Python]
# Guidance function example in the context of grabbing a mug
def guidance_code(state,
    hidden_state={"mug_reached": False,"mug_grabbed":False}):
    #available grounding functions
    #x,y,z = get_pose("mug")
    #h,w,d = get_size("mug")
    #rx,ry,rz = get_orientation("mug")
    ...
    return score, new_hidden_state
\end{lstlisting}
The idea of using abstraction in a hidden state has proven to significantly improve the guidance performance, allowing the guidance functions to keep track of the task progress and adapt the guidance to longer horizon tasks. The guided policy can thus be written as: %
\begin{equation}
    \pi_{guided}(a_{g,t} | o_t, s_t, h_t) = \pi(a_{\pi,t} | o_t, s_t) * G( a_{\pi,t} | o_t, s_{t+1}', h_t).
\end{equation}
The complete guidance procedure is summarized in~Algorithm~\ref{alg:guidance_procedure}.  Note that we refer to the self-orchestrated conversation between the agents which yields the guidance code as the function \texttt{Generate\_Guidance\_Function}. For a closer look at the chain of thought employed by each agent please refer to \Cref{sec:app:prompts}.

\begin{algorithm}[th]

    \caption{Guidance procedure of \ourmethod}
    \label{alg:guidance_procedure}
    \hspace*{\algorithmicindent} \textbf{Input} \\
    \hspace*{\algorithmicindent}$\pi$: Base policy\\
    \hspace*{\algorithmicindent}$\mathcal{D}$: Dynamics Model\\
    \hspace*{\algorithmicindent}$\text{env}$: Environment \\
    \vspace{-0.3cm}
\begin{algorithmic}[1]
\For{each episode}
    \State $o_t$, $s_t \gets$ env.init \CommentColored{observation and initial state}
    \State $G \gets$ Generate\_Guidance\_Function($o_t$, $s_t$)
    \State $h_t \gets \text{Get\_Hidden\_State}(G(o_t, s_t))$ \hspace{-0.3cm}
    
    \For{each time step $t$}

            \State $\mathcal{A}_{\pi,t} \gets \{\pi(o_t,s_t)^{i}\}_{i=0}^n$ 
            \CommentColored{Sample $n$ actions from}\vspace{-0.2em}
            \CommentColoredR{the policy}
            
            \State $\pi_{t} \gets \pi(\mathcal{A}_{\pi,t}| o_t,s_t)$ \CommentColored{Get action probabilities}
            
            \State $\mathcal{S}_{\pi,t} \gets \mathcal{D}(s_t, \mathcal{A}_{\pi,t})$ \CommentColored{Infer possible future states}\\
            \State $G_{\pi,t} \gets G(o_t, \mathcal{S}_{\pi,t}, h_t)$ \CommentColored{Compute the guidance}\vspace{-0.2em}
            \CommentColoredR{for the sampled possible future states}

            \State Normalize $G_{\pi,t}$

            \State $\pi_{\text{guided},t} \gets \pi_{t} * G_{\pi,t}$ \CommentColored{Combine distributions}

            \State $a_t \gets \mathcal{A}_{\pi,t}\left[\text{\small{argmax}}(\pi_{\text{guided},t})\right]$ \CommentColored{Select the best}\vspace{-0.2em}
            \CommentColoredR{action}
            

        \State $o_t$, $s_t \gets$ env.step($a_t$) \CommentColored{Execute $a_t$, update state}\vspace{-0.2em}
        \CommentColoredR{$s_{t+1}$ and observation $o_{t+1}$}

        \State $h_t \gets \text{Get\_Hidden\_State}(G(o_t, s_t, h_t))$ \CommentColored{Update}\vspace{-0.2em}
        \CommentColoredR{hidden states}
        
    \EndFor
\EndFor
\end{algorithmic}
\end{algorithm}

\def\colsize{1.7cm}
\definecolor{darkGreen}{rgb}{0.20, 0.60, 0.20}
\newcommand{\imp}[1]{\textcolor{darkGreen}{(+#1)}}
\newcommand{\nimp}[1]{\textcolor{gray}{}}

\begin{table*}[!t]
    \centering
    \caption{Performance improvement on the RL-Bench \citep{james2020rlbench} benchmark, by applying 5 iterations of guidance improvement over unsuccessful rollouts.}
    \resizebox{\textwidth}{!}{
    \begin{tabular}{p{3cm}|p{\colsize}p{\colsize}p{\colsize}p{\colsize}p{\colsize}p{\colsize}p{\colsize}p{\colsize}p{\colsize}p{\colsize}|p{1.9cm}}
        \toprule
        
        \textbf{Model} & turn tap & open drawer & sweep to dustpan of size & meat off grill & slide block to color target & push buttons & reach and drag & close jar & put item in drawer & stack blocks & \textbf{Avg. Success}\\
        
        \midrule
    \midrule
    Act3D 25 demos/task  & 76 & 76 & 96 & 64 & 92 & 84 & 96 & 48 & 60 & 0 & 69.2\\
    \midrule

     +1\% guidance & 80 \imp{4} & 96 \imp{20} & 96 \nimp{0} & 84 \imp{20} & 92 \nimp{0} & 84 \nimp{0} & 100 \imp{4} & 84 \imp{36} & 80 \imp{20} & 8 \imp{8} & \textbf{80.4 \imp{11.2}} \\
     +10\% guidance & 88 \imp{12} & 100 \imp{24} & 96 \nimp{0} & 88 \imp{24} & 92 \nimp{0} & 84 \nimp{0} & 100 \imp{4} & 60 \imp{12} & 80 \imp{20} & 0 \nimp{0} & 78.8 \imp{9.6} \\
    \midrule
    \midrule

    Act3D 10 demos/task  & 32 & 60 & 84 & 16 & 60 & 72 & 68 & 32 & 44 & 8 & 47.6\\
    \midrule

     +1\% guidance & 44 \imp{12} & 88 \imp{28} & 88 \imp{4} & 24 \imp{8} & 68 \imp{8} & 72 \nimp{0} & 76 \imp{8} & 52 \imp{20} & 60 \imp{16} & 8 \nimp{0} & \textbf{58 \imp{10.4}} \\
     +10\% guidance & 44 \imp{12} & 64 \imp{4} & 84 \nimp{0} & 20 \imp{4} & 68 \imp{8} & 76 \imp{4} & 76 \imp{8} & 40 \imp{8} & 56 \imp{12} & 8 \nimp{0} & 53.6 \imp{6} \\
    \midrule
    \midrule

    Act3D 5 demos/task  & 24 & 0 & 84 & 4 & 8 & 32 & 8 & 8 & 12 & 0 & 18\\
    \midrule

     +1\% guidance & 48 \imp{24} & 16 \imp{16} & 84 \nimp{0} & 8 \imp{4} & 12 \imp{4} & 40 \imp{8} & 24 \imp{16} & 20 \imp{12} & 20 \imp{8} & 0 \nimp{0} & \textbf{27.2 \imp{9.2}} \\
     +10\% guidance & 24 \nimp{0} & 0 \nimp{0} & 84 \nimp{0} & 8 \imp{4} & 12 \imp{4} & 44 \imp{12} & 20 \imp{12} & 8 \nimp{0} & 20 \imp{8} & 0 \nimp{0} & 22 \imp{4} \\
    \midrule
    \midrule
    Diffuser actor 5 demos/ task & 24 & 64 & 40 & 28 & 44 & 68 & 40 & 24 & 44 & 0 & 37.6\\
    \midrule

    +1\% guidance & 40 \imp{16} & 92 \imp{28} & 64 \imp{24} & 40 \imp{12} & 44 \nimp{0} & 68 \nimp{0} & 52 \imp{12} & 24 \nimp{0} & 84 \imp{40} & 0 \nimp{0} & \textbf{50.8 \imp{13.2}}\\
    +10\% guidance & 40 \imp{16} & 84 \imp{20} & 52 \imp{12} & 28 \nimp{0} & 52 \imp{8} & 68 \nimp{0} & 48 \imp{8} & 32 \imp{8} & 76 \imp{32} & 0 \nimp{0} & 48 \imp{10.4}\\
    \bottomrule
\end{tabular}
}
\label{tab:rlbench_results}
\end{table*}

\subsection{Guidance and policy integration}
\label{subsec:guidance_policy_integration}

Aiming to guide a wide range of policies, our framework is designed to work both with continuous and discrete action spaces. In this section, we discuss the operation of combining the guidance function with the base policy's action distributions. Furthermore, we discuss how deterministic regression models can be adapted to work with our framework.

\para{Action-space Adaptation.} We assume the availability of a dynamics model $\mathcal{D}$ that can forecast possible future states of the robot given a possible action $a_{\pi,t}'$. In the manipulation domain, a dynamics model is often available in the form of a forward kinematics model, a learned dynamics model, or a simulator. Oftentimes, the action space $A$ of policies them-self is the same as the robot's state $S$ either being or joint angles of the robot or the gripper's end-effector pose. For the last cases, where both the action and state space are expressed in $SE(3)$ integrating the guidance function with a base policy would only require a multiplication of the guidance scores with the action probabilities of the base policy. In other scenarios, adapting the robot's action and state space to match the representation of the visual cues (position, orientation, and size) would be required.

Considering the visual grounding, the action space and the state space share the same representation ($SE(3)$), the operation to combine the guidance function with the base policy can be expressed as an element-wise weighted average: %
\begin{equation}
    \label{eq:weighted_average}
    \pi_{guided} = (1- \alpha) \pi \cdot \alpha G,
\end{equation}
where $\alpha \in [0, 1]$ represents the percentage of guidance applied with respect to the base-policies distribution and is here denoted as \textit{guidance factor}.




\para{Adaptation of Regression Policies.} To properly leverage the high-level guidance expressed in the guidance functions and the low-level capabilities of the base policy, it is desired that the policy's action space be expressed as a distribution. In the case of regression policies that do not provide uncertainty estimates, several strategies can be employed to infer the action distribution. One common approach is to assume a Gaussian distribution centered at the predicted value and compute the variance using ensembles of models trained with different initialization, different data samples, or different dropout seeds or different checkpoint stages \citep{abdar2021review}. Other strategies to infer the distributions of the model include using bootstrapping, Bayesian neural networks, or using Mixture of Gaussians \citep{mena2021survey}.







%% file: inputs/4_experiments_results.tex
\section{Experiments \& Results}
\label{sec:experiments}

\subsection{Experimental Setup}

\para{Task Definitions:} We demonstrate the efficacy of \ourmethod, in simulation on the RL-Bench benchmark \cite{james2020rlbench} and on two challenging real-world tasks. For the real-world setup, we use the UFACTORY Lite 6 robot arm as the robotic agent and, as the end-effector, we use the included UFACTORY Gripper Lite, a simple binary gripper. The arm is mounted on a workbench. For perception, we use a calibrated RGB-D Camera, specifically the Intel RealSense Depth Camera D435i. All experiments were conducted on a desktop machine with a single NVIDIA RTX 4080 GPU, 64GB of RAM, and a AMD Ryzen 7 8700G CPU.

\begin{itemize}[noitemsep, left=0pt, topsep=0pt]

    \item \textbf{\textit{(Sim-to-real)}: Button-pressing:} Here we employ the diffuser policy to perform the button pressing tasks; to make the task more challenging we introduce clutter in the environment, as well as employ out-of-distribution props. The intent is to measure the ability of \ourmethod to mitigate the sim-to-real gap and remain robust to distractor objects.

    \item \textbf{\textit{(Sim)}: RL-Bench:} We consider 10 tasks on the RL-Bench benchmark  \cite{james2020rlbench} using a single RGB-D camera input, as described in the \textit{GNFactor} setup \citep{ze2024gnfactormultitaskrealrobot}.

    \item \textbf{\textit{(Sim)}: RL-Bench, learning from scratch:} Aiming to explore the capabilities of \ourmethod on learning new skills from scratch, we selected 4 challenging tasks from the RL-Bench benchmark:
    \textit{turn tap}, \textit{push buttons}, \textit{slide block to color target}, \textit{reach and drag}. 

    \item \textbf{\textit{(Real)}: Sim-to-real policy adaptation:} We want to evaluate the capabilities of \ourmethod on reducing the sim-to-real gap by guiding policies that were pre-trained in simulation but are deployed to real-world tasks.
    
    \item \textbf{\textit{(Real)}: Reach for chess piece:} Given a cluttered scene with many similar objects, we want to evaluate if the multi-granular perception framework can effectively guide the agent to identify and reach for the appropriate target. We implement this perceptual grounding and reaching task on a standard chessboard, where the agent must identify and reach for one of the chess pieces specified by natural language instruction.
\end{itemize}

\para{Base Policies:} We evaluate the effectiveness of our guidance framework using different base policies, \textit{Act3D} \citep{gervet2023act3d}, \textit{3D Diffuser Actor} \citep{ke20243d}, and a RandomPolicy. All policies plan in a continuous space of translations, rotations, and gripper state ($SE(3) \times \mathbb{R}$), however they utilize different inference strategies:

\begin{itemize}[noitemsep, left=0pt, topsep=0pt]
\item \textbf{\textit{Act3D}} samples waypoints in the Cartesian space ($\mathbb{R}^3$) and predicts the orientation and gripper state for the best scoring sampled waypoint, combining a classification and regression strategies into a single policy.
\item \textbf{\textit{3D Diffuser Actor}}, on the other hand, uses a diffusion model to compute the target waypoints and infers the orientation and gripper state from the single forecast waypoint, thus tackling the problem as a single regression task.
\item \textbf{\textit{RandomPolicy}} denotes any of the former frameworks that has not been trained for a specific task, therefore the weights are randomly initialized. 
\end{itemize}


The fundamentally different types of policies' outputs make them a great use case for our policy guidance framework. Furthermore, the common representation of the action and state spaces of both policies ($SE(3) \times \mathbb{R}$) provides a straightforward integration with our grounding models.

As described in Section \ref{subsec:guidance_policy_integration}, the regression components of the policies require an adaptation to transform the single predictions of the model into a distribution over the action space. For the sake of simplicity, we assume a Gaussian distribution over the action space, with the mean centered on the predicted values and the standard deviation fixed on a constant value. The outputs of the classification component of Act3D (waypoint positions) were directly considered as samples of a distribution over the Cartesian space ($\mathbb{R}^3$).

The integration of base policies with the guidance distributions was performed by applying a weighted average parameterized by $\alpha$ as shown in Equation \ref{eq:weighted_average}.


\subsection{Experimental Evaluation}

Our experimental evaluation aims to address the following questions: (1) Does \ourmethod enable policies to maintain or improve performance on tasks in out-of-distribution settings? (2) Does \ourmethod enable policies to bridge the sim-to-real gap for deployment on real-world tasks? (3) Does \ourmethod improve the performance of pre-trained base policies on specific robotics tasks and environments without additional human demonstrations? (4) Does the proposed multi-granular perception capabilities effectively guide the policy in challenging cluttered environments? (5) Does \ourmethod enable policies to learn new skills from scratch? (6) What is the effect of guidance on expert versus untrained policies?


\begin{figure*}[ht]
    \centering
    \includegraphics[width=0.8\textwidth]{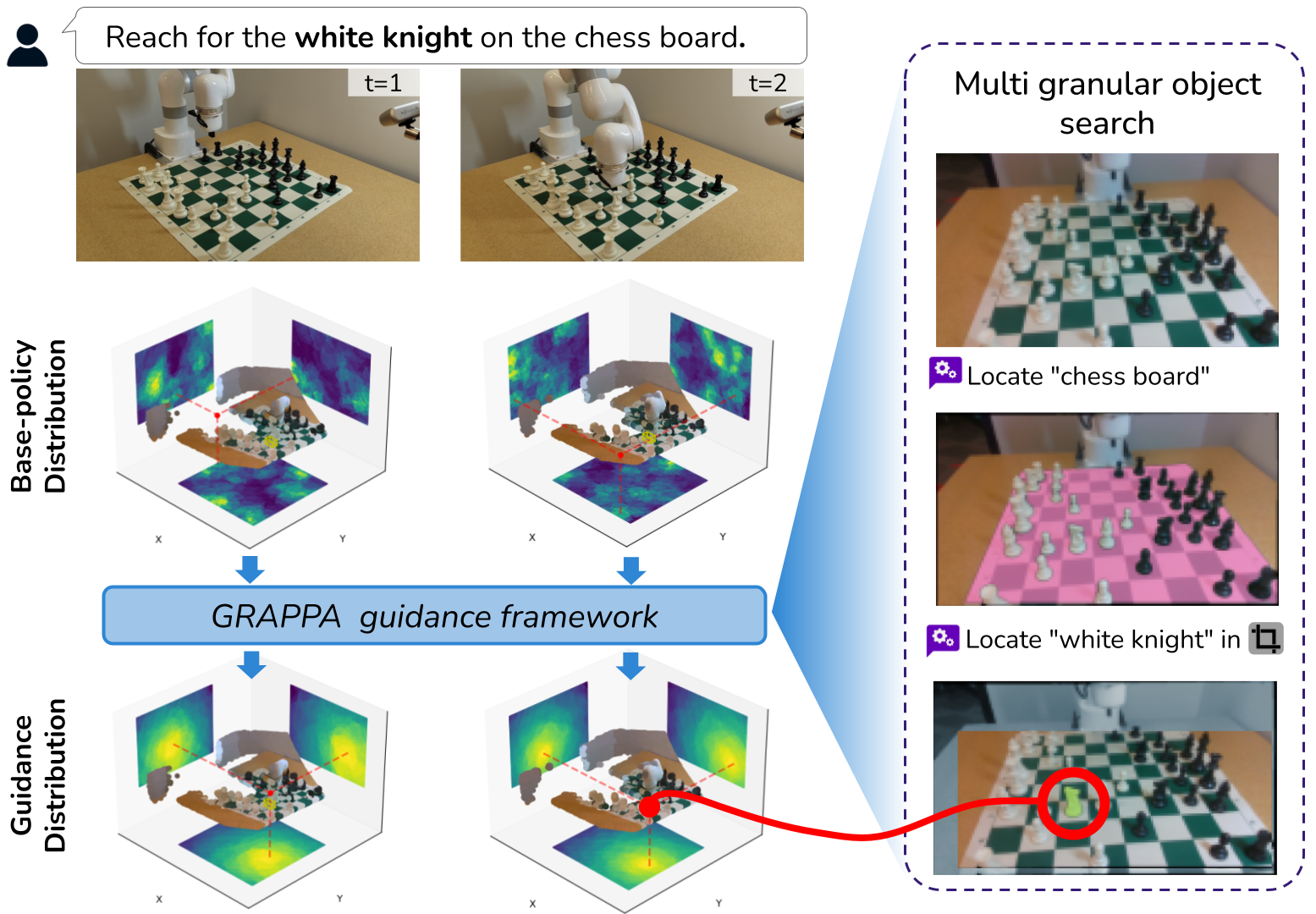}
    \caption{Real-world results for learning skills from scratch on the UFactory xArm Lite6 chess task.  The top row shows an external view of the robot performing the tasks. The second row depicts the action heat map given by the random diffuser policy at the first and last time step. The bottom row depicts the corresponding heat maps generated after applying the guidance. We show it can successfully guide the action towards the desired object. On the right, we show a breakdown of the multi-granular search performed by other \textbf{grounding agent} to locate the white knight; we disambiguate the scene by searching in parent objects and constraining the search to semantically relevant areas. }
    \label{fig:real_results_chess}
\end{figure*}

\para{Does \ourmethod enable policies to bridge the sim-to-real gap for deployment on real-world tasks and in out-of-distribution settings?} We first train a \textit{3D Diffuser Actor} policy with 10 tasks with 100 demonstrations each with a single RGB-D camera setup (\textit{GNFactor} \citep{ze2024gnfactormultitaskrealrobot} setting), using exclusively simulation-based demonstrations. Note that this policy was chosen as it displayed stronger overall performance in the simulation benchmark. We roll out this policy in the real world in the button-pressing task in a cluttered environment, as depicted by Figure \ref{fig:close_up_guidance}. To facilitate sim-to-real integration, we adapted the dimensions, scale, and alignment of the real-world workspace to match the simulation data, and positioned the RGB-D camera (Intel RealSense) in a similar placement as the camera used to train the base policy. We made sure to first validate the sim-to-real setup by performing simple tasks in uncluttered settings as shown in Appendix Table~\ref{tab:simple_sim2real}.

\begin{table}[h]
\centering
\caption{Real-world performance improvement in a sim-to-real setting. \textit{Diffuser actor} policy 
trained on 10 tasks in simulation (\textit{GNFactor} \citep{ze2024gnfactormultitaskrealrobot} setup: 100 demos/task) and evaluated on the tasks \textit{``Push buttons"} with different guidance levels. \textcolor{red}{\textbf{Red}} cells refer to failures due to timeouts in task execution. \textcolor{orange}{\textbf{Orange}} cells refer to errors in perception. \textcolor{green}{\textbf{Green}} cells refer to successes.}
\label{tab:real_quantitative}
\resizebox{\columnwidth}{!}{
\begin{tabular}{lrrcccccc}
\toprule
 \multicolumn{1}{l}{Policy}                                      & Task completion Success Rate [\%] & \multicolumn{6}{l}{Error Breakdown} \\
 \midrule
 \midrule
 \multicolumn{1}{l}{\textit{\textbf{Naive Sim2Real}}}    & 0.0  & \cellcolor{red} & \cellcolor{red}   & \cellcolor{red} & \cellcolor{red}   & \cellcolor{red} & \cellcolor{red}  \\
 \midrule
 +15\% guidance                                            & 66.7 & \cellcolor{green} & \cellcolor{green} & \cellcolor{green} & \cellcolor{green} & \cellcolor{orange} & \cellcolor{red}  \\
 +50\% guidance                                            & 50.0 & \cellcolor{green} & \cellcolor{green} & \cellcolor{green} & \cellcolor{orange} & \cellcolor{red} & \cellcolor{red} \\
 +75\% guidance                                            & 100 & \cellcolor{green} & \cellcolor{green} & \cellcolor{green} & \cellcolor{green} & \cellcolor{green} & \cellcolor{green} \\
 \midrule
 \midrule

 \multicolumn{1}{l}{\textit{\textbf{Finetuned Baseline} (10 real-world demos)}} & 33.0 & \cellcolor{green} & \cellcolor{green} & \cellcolor{red} & \cellcolor{red} & \cellcolor{red} & \cellcolor{red} \\
 \midrule
 +15\% guidance                                            & 50.0 & \cellcolor{green} & \cellcolor{green} & \cellcolor{green} & \cellcolor{red} & \cellcolor{red} & \cellcolor{red} \\
 +50\% guidance                                            & 16.7 &  \cellcolor{green} & \cellcolor{orange} & \cellcolor{orange} & \cellcolor{red} & \cellcolor{red} & \cellcolor{red}  \\
 +75\% guidance                                            & 100 & \cellcolor{green} & \cellcolor{green} & \cellcolor{green} & \cellcolor{green} & \cellcolor{green}  & \cellcolor{green}  \\
 \bottomrule
\end{tabular}
}
\end{table}

As depicted by Table \ref{tab:real_quantitative}, the cluttered scenario in Figure \ref{fig:close_up_guidance} has shown to be out-of-distribution scenarios for the base policy that was solely trained in simulation (Naive Sim2Real). Fine-tuning the policy with 10 real-world demonstrations of pressing buttons in uncluttered environments has proven to slightly increase the performance of the policy. These base policies, combined with our guidance framework, achieve a higher success rate in 5 out of the 6 scenarios with guidance. In the only exception case (finetuned baseline + 50\%), the base perception models failed to detect and keep track of the target button during the rollout---leading to a detrimental guidance of the policy. For a more detailed breakdown of failure cases, we refer the reader to \Cref{sec::app::failure_analysis}.

\begin{figure}[H]
\centering
\includegraphics[width=\linewidth]{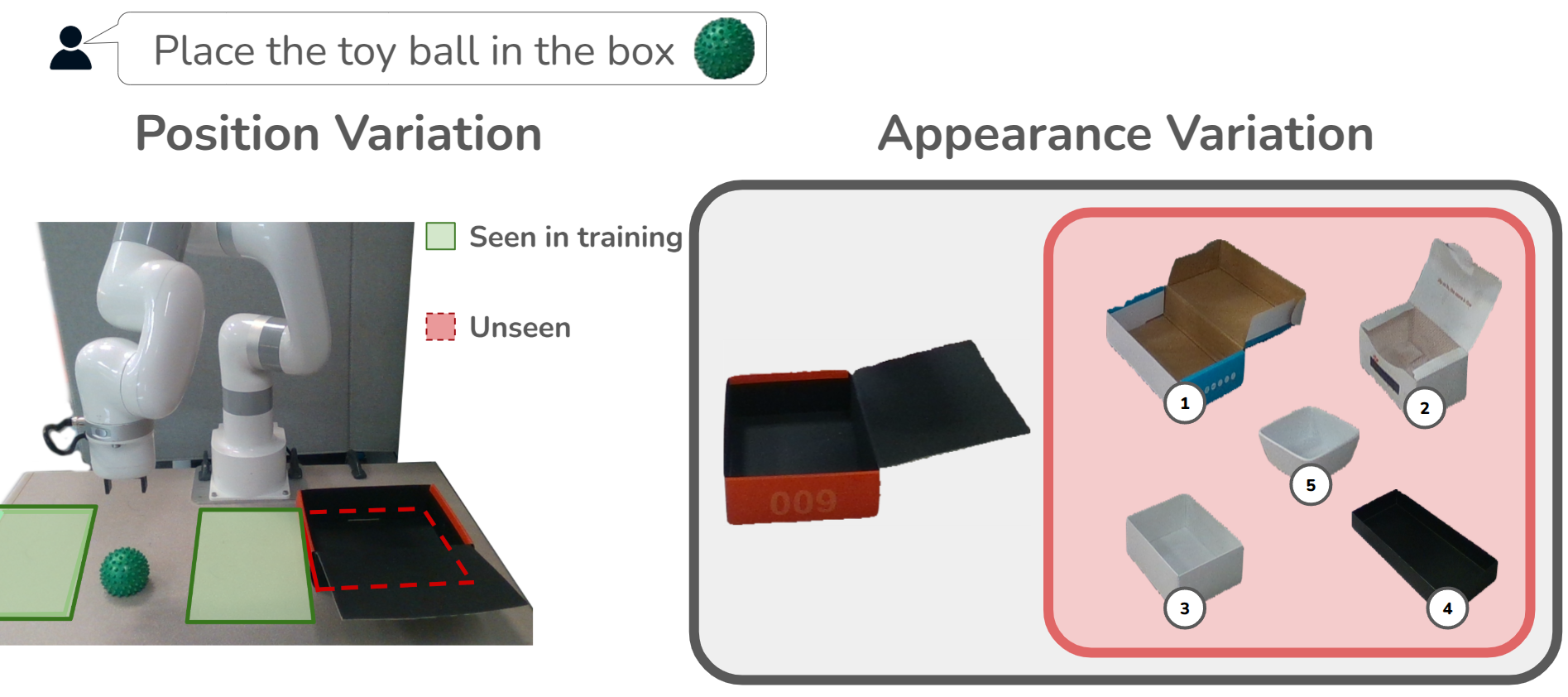}
\caption{\ourmethod guiding the base policy for out-of-distribution cases. The task involves grasping a deformable toy ball and placing it inside a box.}
\label{fig:generalization_exp}
\end{figure}

\begin{table}[!h]
\centering
\vspace{-5.7mm}
\caption{\ourmethod guiding the base policy for out-of-distribution cases, illustrated in Figure \ref{fig:generalization_exp}.}
\label{tab:generalization_exp}
\begin{tabular}{l|c| |ccccc|c}
    \toprule
     & \multicolumn{1}{|c||}{\textbf{Position}} &  \multicolumn{5}{c|}{\textbf{Appearance}}  & \\
    \textbf{Baseline} & \multicolumn{1}{|c||}{\textbf{SR}} & \multicolumn{1}{c}{\textbf{1}} & \textbf{2} & \textbf{3} & \textbf{4} & \textbf{5} & \textbf{SR} \\
    \midrule
    3D Actor Diffuser & 0/5 & \xmark & \xmark & \xmark & \xmark  & \xmark & 0/5  \\
    + \ourmethod(50\%)  & 3/5& \cmark & \xmark & \cmark & \cmark  & \cmark & 4/5 \\
    \bottomrule
\end{tabular}

\end{table}

\begin{figure}[t]
    \centering
    \includegraphics[width=\linewidth]{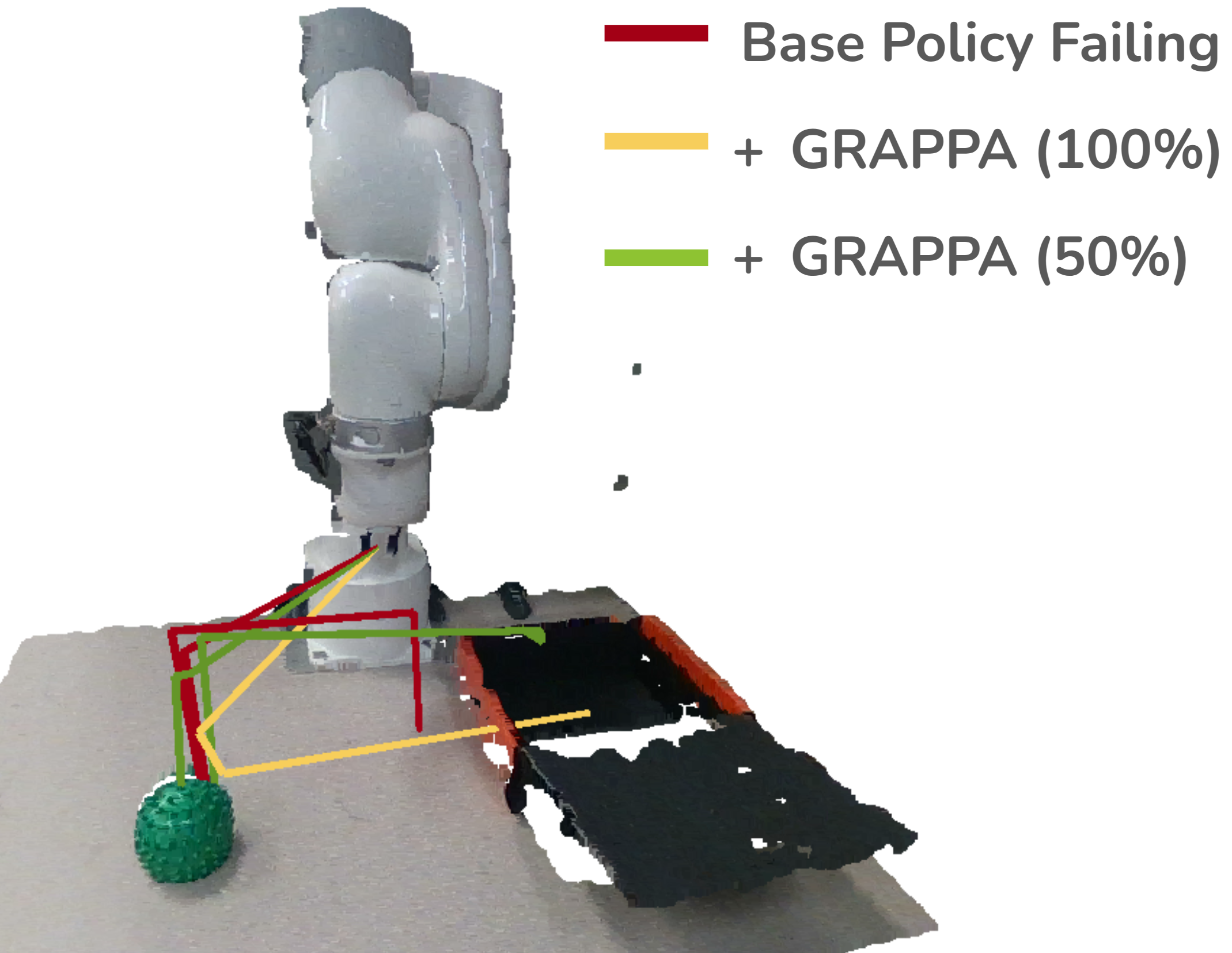}
    \caption{Illustration of the effect of different guidance percentages on a failure case of the base policy. In red we show the base policy failing in an out-of-distribution scenario; with 100\% of guidance (yellow), the end position is successfully above the box, but it has \textit{lost low-level notions}. By balancing both with intermediate guidance (50\%) shown in green, we can complete the task.}
    \label{fig:guidance_comparison}
    \vspace{-0.5cm}
\end{figure}

A known limitation of diffusion-based policies is their struggle to generalize beyond their training set. To test the ability of \ourmethod to remedy this type of situation, we again take a policy trained in simulation and finetune on 15 demonstrations of a pick-and-place task in the real world. We assess across two axes of generalization: 1) position generalization, where the objects are the same as used in the training demonstrations, but the starting positions are different; and 2) appearance generalization, where we vary the type of object used, while keeping the same semantic class of objects. \Cref{tab:generalization_exp} showcases the qualitative results for 10 rollouts: we see that in both types of generalization, the 3D Actor Diffuser is unable to complete the task in any trial, while \ourmethod using 50\% of guidance succeeds in $3$ out of $5$ cases for position generalization and $4$ out of $5$ for appearance generalization.

Across both real-world tasks, a higher percentage of guidance is necessary compared to the simulation results seen in \Cref{tab:rlbench_results}. This hints at two practical properties of \ourmethod's guidance effect. First, there must be a trade-off between high-level guidance (provided by \ourmethod) and low-level control (provided by the base policy), in more complex tasks, elevated guidance values disrupt the robot's low-level motions, leading to some failure cases. We compare trajectories of the failing base policy versus \ourmethod-guided in \Cref{fig:guidance_comparison}. Second, the optimal guidance level is also dependent on the performance of the base policy itself. Notably, while the simulation-based policy has already achieved a good performance on the test set, it is unable to deal with out-of-distribution cases and requires much more correction in the real-world setting.

\para{Does \ourmethod improve the performance of pre-trained base policies on specific robotics tasks and environments without additional human demonstrations?} We first assess the effect of the proposed guidance on the \textit{Act3D} and \textit{3D Diffuser Actor} baselines following the \textit{GNFactor} \citep{ze2024gnfactormultitaskrealrobot} setup, which consists of a single RGB-D camera and table-top manipulator performing 10 challenging tasks with 25 variations each. Guidance is iteratively generated for the failure cases. For the failed rollouts, our policy improvement framework ran for 5 iterations. As displayed by Table \ref{tab:rlbench_results}, the framework was able to improve the success rate of the base policy on most of the tasks, with the best results achieved by using 1\% guidance. The low amount of guidance has shown to be enough to bend the action distribution to the desired direction, while still preserving the low-level nuances captured by the base policy. This suggests that \ourmethod is capable of improving base policies by adding abstract understanding and grounding of the desired task, while preserving the low-level movement profiles captured by the original policies. We provide additional rollouts of \ourmethod guidance correcting a previously-failing base policy, in Appendix Figure~\ref{fig:rollouts}.


\para{Does the proposed multi-granular perception capability effectively guide the policy in challenging, cluttered environments?} In real-world experiments, we qualitatively demonstrate the fine-grained detection capabilities of \ourmethod by tasking it with reaching for a white knight chess piece in a cluttered chess board. Figure \ref{fig:real_results_chess} shows the roll-out of the first guidance iteration over an untrained policy, displaying the initial and final time steps of the task. The accompanying heatmaps illustrate the distributions of the original untrained policy (Diffuser heatmaps) and the guided policy (Guidance heatmaps). Additionally, the multi-granular search results highlight the steps taken by the grounding agent to locate the target piece. After initially failing to detect the white knight directly, the agent successfully identifies the chessboard and then focuses within that region to locate the target piece. These findings demonstrate that \ourmethod effectively leverages a semantic understanding of scene components to guide the policy towards successful task execution.

\begin{figure}[h]
     \centering
     \includegraphics[width=\columnwidth]{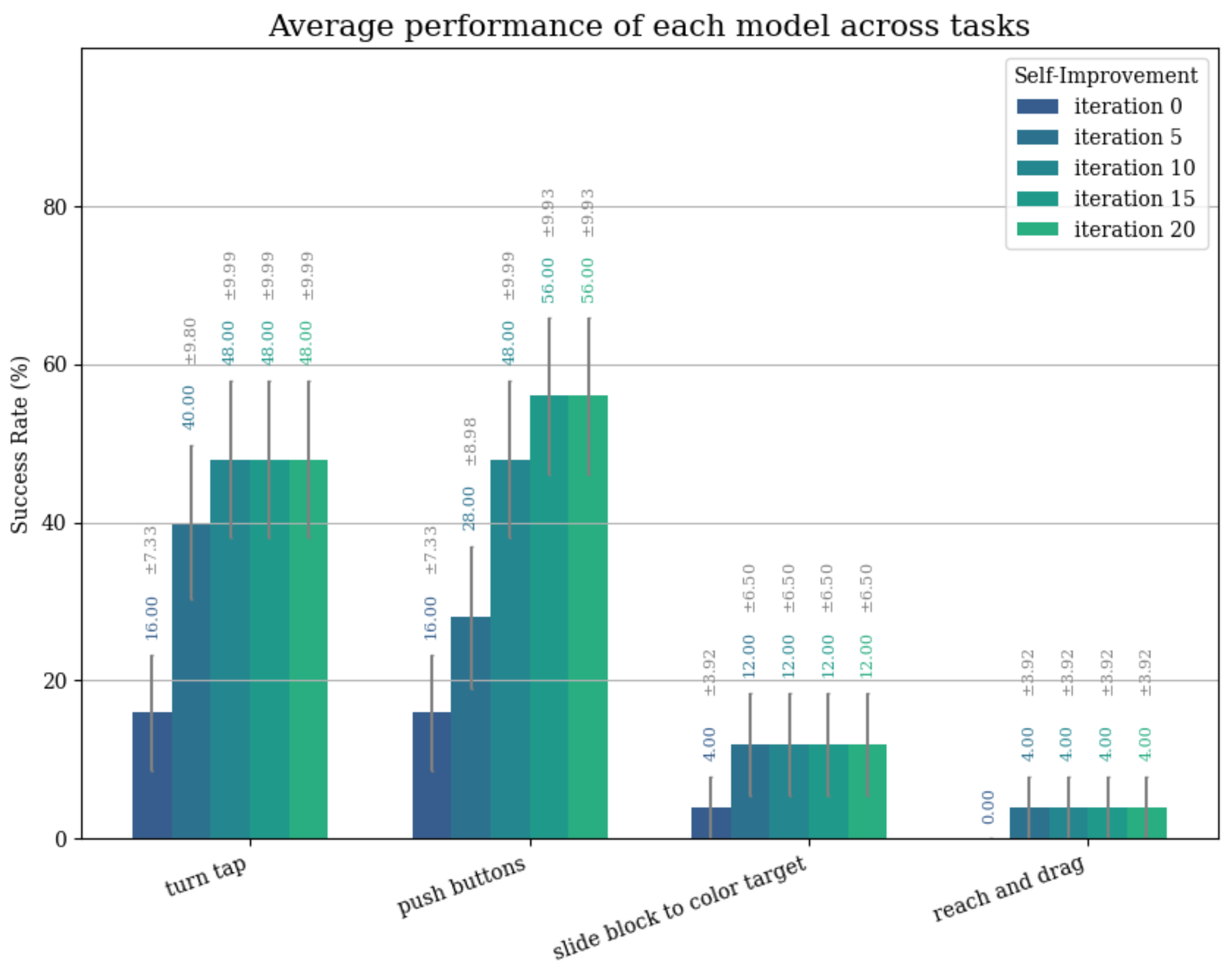}
     \caption{Performance of our framework on learning new skills from scratch (guidance over untrained policies), and iteratively improving the guidance functions generated.}
     \label{fig:iteration_results}
 \end{figure}

\para{Does \ourmethod enable policies to learn new skills from scratch?} We evaluated the performance of the framework on learning new skills from scratch on 4 tasks of the RL-Bench benchmark: \textit{turn tap}, \textit{push buttons}, \textit{slide block to color target}, \textit{reach and drag}. In this setup, we initialized the Act3D policy with random weights and applied 100\% of guidance ($\alpha = 1.0$) over the policy for the x,y, and z components. Only leveraging the waypoint sampling mechanism of Act3D and overwriting its distribution with the values queried from the guidance functions generated. The results show that the framework is capable of learning new skills from scratch, achieving a higher success rate than the base policy pre-trained on 5 demonstrations/tasks for the tasks ``turn tap" and ``push buttons" tasks. When utilizing only the untrained Act3D policy (without guidance) the policy achieved 0 success rate on the tasks. Figure \ref{fig:iteration_results} demonstrates the iterative improvement of our guidance framework, wherein the guidance code generated for each failed rollouts from the previous iteration is iteratively updated. Policy rollouts are provided in Appendix Figure \ref{fig:rollouts}. 

It is worth mentioning that a few variations of the simulated tasks ``turn tap" and ``reach and drag", which seemly would require a precise orientation control of the manipulator, could be solved by guiding only the Cartesian components of the police's output. For these variations, a qualitative analysis shows that successful roll-outs could be achieved by tapping the end-effector on the target objects.

\begin{figure}[ht]
    \centering
    \includegraphics[width=\columnwidth]{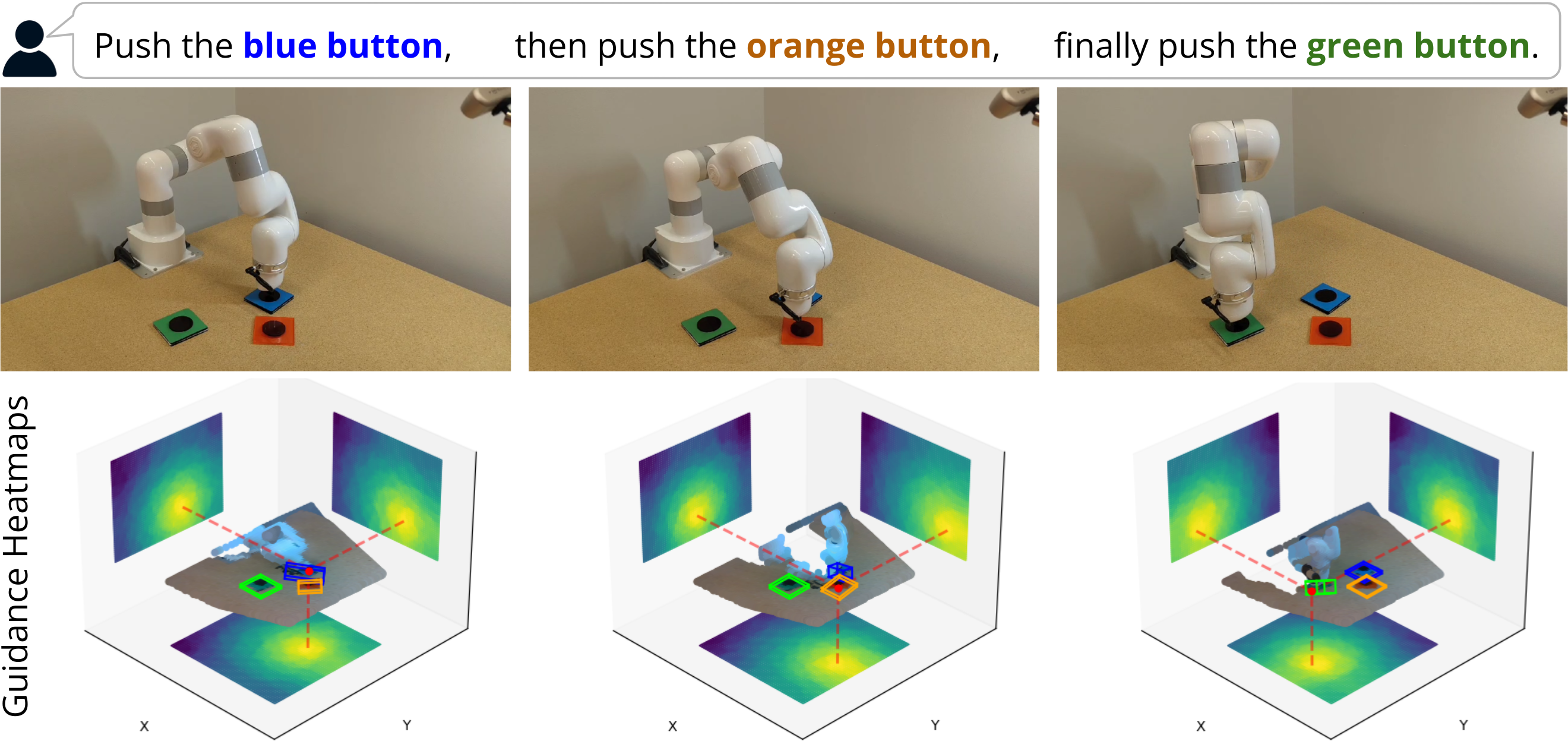}
    \caption{Real-world results for learning skills from scratch on the Lite6 on the multiple button press task. Each column represents a keyframe in the task rollout. The first row shows a third-person view of the robot's movement and the tabletop setting. The second shows the corresponding action distribution over the space generated by the guidance code, the red dot indicates the target waypoint for the end effector. We demonstrate that \ourmethod can guide a random base policy to successfully perform the desired task. }
    \label{fig:real_results}
\end{figure}

We perform a similar experiment in the real-world settings. Here, we run the framework only relying on the action distribution given by the guidance code ($\alpha = 1.0$), using a RandomPolicy as the base policy. We first consider the task of pressing colored buttons in a given sequence; using toy buttons made out of acrylic and paper as a prop. Figure \ref{fig:real_results} shows the roll-out corresponding to the first iteration for this task, along with heat maps depicting a projection of the output action distribution around the point of maximum. In this zero-shot scenario, \ourmethod has proven to correctly guide the robot to the desired objects preserving the prompted order; however, it struggles to capture low-level nuances of the movement, such as the appropriate pressing force and proper approach of the buttons. A simulated version of this experiment is shown in Appendix Figure~\ref{fig:rollouts}, demonstrating how combining the guidance with a pre-trained policy can mitigate this behavior.

\para{What is the effect of guidance on expert versus untrained policies?} As discussed previously, \ourmethod can learn tasks from scratch using a random base policy with 100\% of guidance $(\alpha)$. Given that we are not affecting the policy, the product of the iterative learning from scratch is the generated guidance script which captures a high-level understanding of the task, e.g., spatial relationships, and task completion criteria, among others. We can qualitatively see the effect of the learning process by comparing the guidance scripts for different iterations of the same task. Appendix \ref{subsec:guidance_code_samples} includes two samples of guidance code for the same task but on different iterations. We can see that the code corresponding to the second iteration is very simple and only accounts for Euclidean distance and button order; while by the fifth iteration considerations like orientation come into play.

On the other hand, applying the guidance to an existing expert policy aims to shift the action distribution to account for failure cases, like potentially out-of-distribution scenes. The goal here is to use the gained high-level understanding to aid the policy in task completion. Table \ref{tab:rlbench_results} shows that adding too much weight to the guidance function can yield diminishing returns as it can overpower the nuanced low-level details from the expert policy: notice that the performance gain across the board is bigger using 1\% of guidance.

%% file: inputs/5_discussion.tex
\section{Conclusion}
\label{sec:conclusion}

\para{Summary:} 
In this work, we proposed \ourmethod, a novel framework for the self-improvement of embodied policies. Our self-guidance approach leverages the world knowledge of a group of conversational agents and grounding models to guide policies during deployment. We demonstrated the effectiveness of our approach in autonomously improving manipulation policies and learning new skills from scratch, in simulated RL-bench benchmark tasks and in two challenging real-world tasks. Our results show that the proposed framework is especially effective in improving the following high-level task structures and key steps to solve the task. 
This capability can be well suited for improving pre-trained policies that struggle with long-horizon tasks or for learning new simple skills from scratch.

\para{Limitations:} From an analysis of the guided rollouts, a few of the tasks variations proved challenging for the perception models used by the grounding agent, leading to false positives detections or failure to locate specific objects. This limitation was mainly observed in simulation task, were the graphics object representations, even though simplified, do not always match the representations used to train the object detection models. This limitation could be addressed by integrating more robust object detection models or verification procedures to ensure the correct detection of objects in the scene. Moreover, occasional inaccuracies on scene understanding by the Visual Language Model (VLM) have been observed, leading  to the generation of inaccurate guidance codes and unexpected behaviors. Even though recent advances in large vision-language models have shown great potential in understanding the underlying dynamics of the world from large-scale internet data, translating this knowledge into out-of-distribution domain, such as robotics, while preventing hallucinations remains an open challenge.

\para{Future Work:} Regarding future works, we think that combining the proposed framework with fine-grained exploration techniques would allow the policy to explore in a targeted manner the low-level details of the task while leveraging the high-level guidance provided by our framework. This may enrich the guidance codes with the necessary low-level details required to perform more complex tasks successfully. 

Furthermore, the guidance function generation could be further improved by composing and adapting from a repository of successful guidance functions from previous experiences. This could be achieved by incorporating Retrieval Augmented Generation (RAG)  \citep{lewis2021retrievalaugmentedgenerationknowledgeintensivenlp} into our multi agent framework. This modification could allow the guidance system to learn new simple skills from scratch by interacting with the environment and leveraging this collected knowledge to guide the policy more effectively.

Aiming to incorporate the knowledge captured by the guidance functions into the base policy, an experience replay and finetuning mechanisms could be incorporated into our current system. This modification could allow the framework to use past guided experiences to improve the base policy in a sample efficient manner. This could be achieved in a targeted matter by leveraging Low Rank Adaptation (LoRA) \citep{hu2021lora}.

\para{Reproducibility Discussion:} Intending to encourage other researchers to build upon the introduced framework, we take steps to ensure the usability and reproducibility of our work. The source code for \ourmethod will be open-sourced and linked to on a project website. We will provide dockerized scripts to facilitate the setup across different development environments. Additionally, in Section \ref{sec:app:prompts} we include the prompts used to configure each agent. The temperature of the model was set as zero to reduce variations in runs, as using fixed seeds for the experiments. More hyperparameter details will be available in the open-sourced repository.

%% file: inputs/6_appendix.tex
\onecolumn
\appendix
\section{Appendix for the paper "\textit{\ourmethod}: Generalizing and Adapting Robot Policies via Online Agentic Guidance"}

\subsection{Failure Analysis}
\label{sec::app::failure_analysis}
\begin{figure}[!th]
    \centering
    \includegraphics[width=0.70\linewidth]{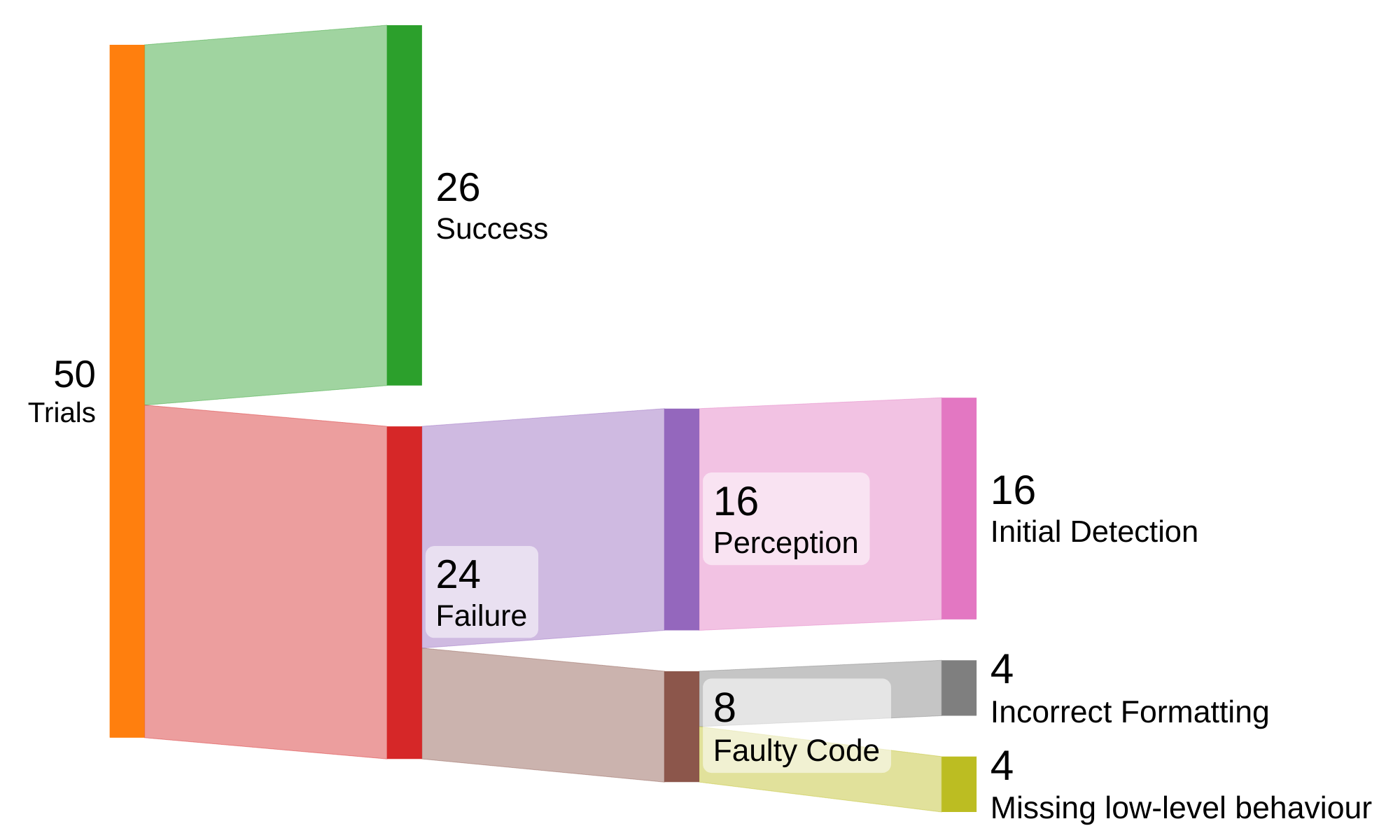}
    \caption{
    Breakdown for failure cases from the learning-from-scratch experiment (push buttons and turn tap), classifying trials by logs, guidance codes, and observed behavior. Note that this analysis is performed on the learning-from-scratch experiment to decouple the errors of \ourmethod from the base policy.}
    \label{fig:failure_analysis}
\end{figure}

\subsection{Agent Prompt Configuration}
\label{sec:app:prompts}

We provide the system prompts used to initialize each one of the agents. Note that for models relying on API calls, we use \texttt{gpt-4o-mini-2024-07-18}. The maximum number of tokens is set to $2000$. 

\begin{figure}[!th]
    \centering
    \includegraphics[width=1.0\linewidth]{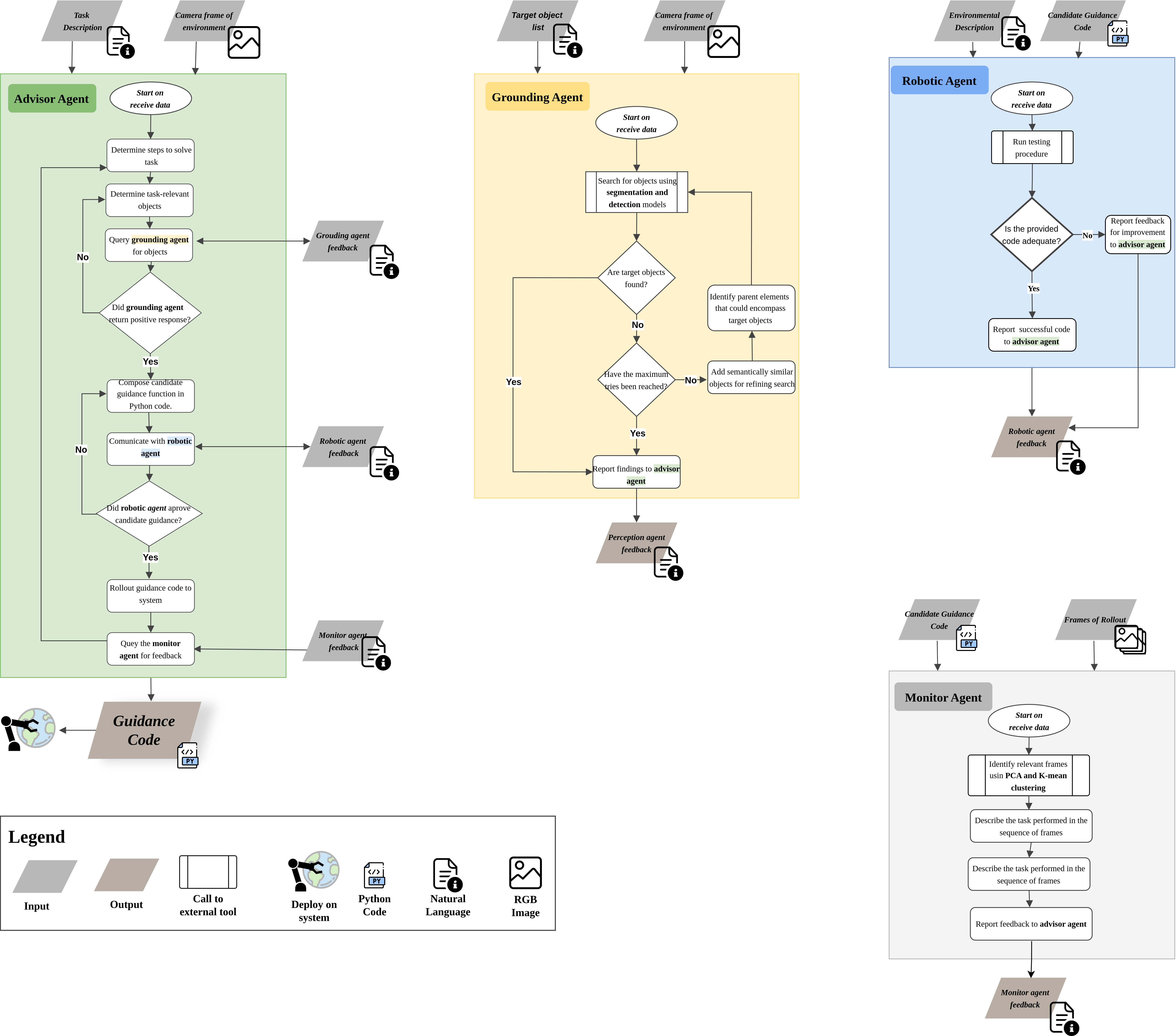}
    \caption{ The agents in the \ourmethod framework are instances of large multimodal models that communicate with each other to produce a final guidance code; leveraging the reasoning capabilities of this type of model. This image exemplifies the chain of thought each agent is encouraged to follow, which in practice is encoded as a natural language prompt shown in \Cref{sec:app:prompts}. The agents can call external tools to aid their analysis such as detection models and a Python interpreter for scrutinizing the code. The advisor agent acts as the main orchestrator, querying the other agents as necessary and generating and refining the guidance code with the provided feedback.}
    \label{fig:enter-label}
\end{figure}

\begin{llmprompt}{roboticblue}{Robotic Agent Prompt}
    You are an AI agent responsible for controlling the learning process of a robot.
    You will receive Python code containing a guidance function that helps the robot with the execution of certain tasks.
    Your job is to analyze the environment and criticize the code provided by checking if the guidance code is correct and makes sense.
    You SHOULD NOT create any code, only analyze the code provided by the supervisor. Attend to the following: \\
      - The score provided by the guidance function is continuous and makes sense. \\
      - The task is being solved correctly. \\
      - The code can be further improved. \\
      - The states of the robot are being correctly expressed. \\
      - The code correctly conveys the steps to solve the task in the correct order.\\
    BE CRITICAL! \\ 
    Make sure that the robot state is expressed as its end-effector position and orientation in the format by using the function test\_guidance\_code\_format().
    If the code is not correct or can be further improved, provide feedback to the supervisor\_agent and ask for a new code.
    Use 'NEXT: supervisor\_agent' at the end of your message to indicate that you are talking with the supervisor\_agent.
    If no code is provided, ask the supervisor\_agent to generate the guidance code.
    If the code received makes sense and is correct, simply output the word 'TERMINATE'.

\end{llmprompt}

\begin{llmprompt}{groundingyellow}{Grounding Agent Prompt}
    You are a perception AI agent whose job is to identify and track objects in an image. 
    You will be provided with an image of the environment and a list of objects that the robot is trying to find (e.g. door, handle, key, etc).
    With that, you can make use of the following function to try to locate the objects in the image: in\_the\_image(image\_path, object\_name, parent\_name) -\textgreater yes/no.
    If the object is not found it might be because the object was too small, too far, or partially occluded, in this case, try to find a broader category that could encompass the object.
    In this case, report the function call used followed by 'NEXT: perception agent' to look for the objects using similar object names or with a parent name that cloud encompasses the object.
    (e.g. first answer: 'in\_the\_image('door handle') -\textgreater no NEXT: perception agent', second answer: 'in\_the\_image('door\_handle', 'door') -\textgreater no NEXT: perception agent', third answer: 'in\_the\_image('handle', 'gate') -\textgreater yes. couldn't find a door handle but found a gate handle NEXT: supervisor\_agent').
    Report back to the supervisor agent in a clear and concise way if the objects were found or not. If an object was found using a parent name, report the parent name and the object name.
    Use 'NEXT: supervisor\_agent' at the end of your message to indicate that you are talking with the supervisor\_agent, or 'NEXT: perception\_agent' to look further for the objects.
\end{llmprompt}

\begin{llmprompt}{advisorgreen}{Advisor Agent Prompt}
    You are a supervisor AI agent whose job is to guide a robot in the execution of a task.
      You will be provided with the name of a task that the robot is trying to learn (e.g. open door) and an image of the environment. With that, you must follow the following steps: \\
      1- determine the key steps to solve the tasks.\\
      2- come up with the names of features or objects in the environment required to solve the task.\\
      3- check if the objects are present in the scene and can be detected by the robot by providing the image to the perception agent and asking the perception agent (e.g. 'Can you find the door handle?' wait for feedback), If the answer goes against of what you expected to repeat the steps 1 to 3.\\
      4- Only proceed to this step after receiving positive feedback on 3. Write a Python code to guide the robot in the execution of the task. The output code needs to have a function that takes the robot's state as input (def guidance(state, previous\_vars={'condition1':False, ...}):), queries the position of different elements in the environment (e.g get\_position('door')) and outputs a continuous score for how close is the robot to completing the task (e.g. the robot is far away from the door the score should be low).\\

      When writing the guidance function, you can make use of the following functions that are already implemented: get\_position(object\_name) -\textgreater [x,y,z], get\_size(object\_name) -\textgreater [height, width, depth], get\_orientation(object\_name) -\textgreater euler angles rotation [rx,ry,rz].
      and any other function that you think is necessary to guide the robot (e.g. numpy, scipy, etc).

      The guidance function must return a score (float) and a vars\_dict (dict). The vars\_dict will be used to store the status of conditions relevant to the task completion. The previous\_vars\_dict input with contain the vars\_dict from the previous iteration.
      The score must be a continuous value having different values for different states of the robot. States slightly closer to the goal should have slightly higher scores.
      The next action of the robot will depend on the score returned by the guidance function when queried for many possible future states. 

      "The state of the robot is a list with 7 elements of the end-effector position, orientation and gripper state [x, y, z, rotation\_x, rotation\_y, rotation\_z, gripper], gripper represents the distance between the two gripper fingers. All distance values are expressed in meters. and the rotation values are expressed in degrees."

      start your code with the following import: 'from motor\_cortex.common.perception\_functions import get\_position, get\_size, get\_orientation'.
      Do not include any example of the guidance function in the code, only the function itself. \\
      \\
      code format example: \\
      ``` \\
      from motor\_cortex.common.perception\_functions import get\_position, get\_size, get\_orientation \\
      \# relevant imports \\
      \# helper functions \\
      def guidance(state, previous\_vars\_dict={'condition1':False, ...}):\\
          \# your code here\\
          return score, vars\_dict\\
      ``` \\
      You are encouraged to break down the task into sub-tasks, and implement helper functions to better organize the code.\\

      You can communicate with a perception\_agent and a robotic\_agent. \\
      Always indicate who you are talking with by adding 'NEXT: perception\_agent' or 'NEXT: robotic\_agent' at the end of your message. 
\end{llmprompt}

\begin{llmprompt}{monitorgray}{Monitor Agent Prompt}
    You will be given a sequence of frames of a robotic manipulator performing a task, and a guidance code used by the robot to perform the task. \\
    Your job is to describe what the sequence of frames captures, and then list how the robot could better perform the task in a simple and concise way. \\
    Do not provide any code, just describe the task and how it could be improved.
\end{llmprompt}




\subsection{Sim-to-Real performance with different buttons}

\def\btcolsize{2cm}

\begin{table}[h]
\centering
\caption{Success rate on the \textit{``push button"} task with out-of-distribution buttons}
\label{tab:simple_sim2real}
\begin{tabular}{l c c c }
\textbf{Policy} & \includegraphics[width=\btcolsize]{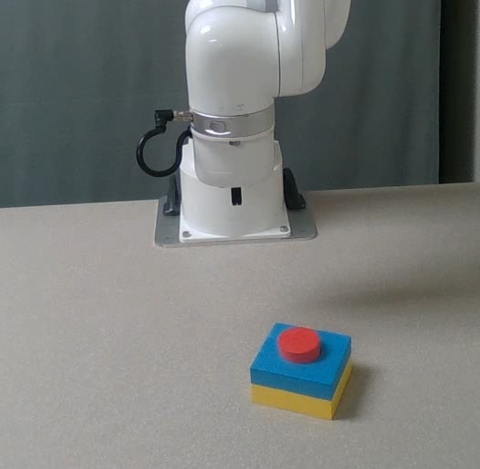} & \includegraphics[width=\btcolsize]{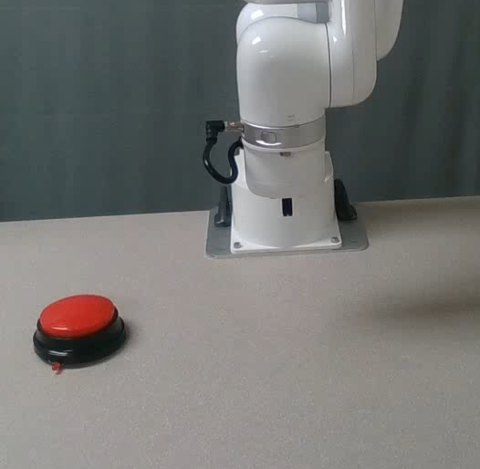} & \includegraphics[width=\btcolsize]{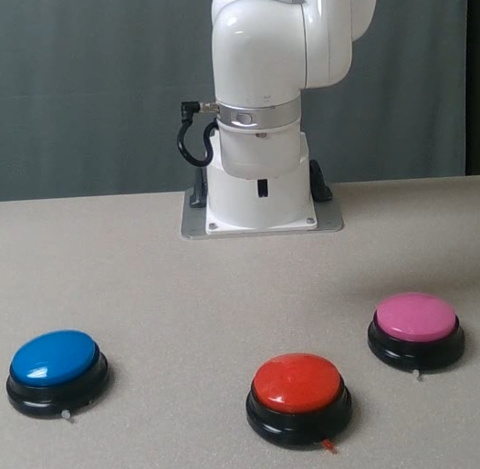}\\
\hline
Naive Sim2Real & 3/6 & 3/6 & 2/6  \\
\bottomrule
\end{tabular}
\end{table}

\subsection{Guidance examples}
\label{subsec:guidance_code_samples}

The output of the guidance function is a Python script containing high-level abstractions and an understanding of the given tasks. Here we include a guidance code example for an order button press.

\begin{lstlisting}[language=Python, caption={Guidance code generated for the button-pushing task in the second iteration.}, label=lst:first_iter]
from grappa.common.perception_functions import get_position, get_size, get_orientation
import numpy as np

# Helper function to check if the button is pressed
def is_button_pressed(button_position, robot_position):
    distance = np.linalg.norm(np.array(button_position) - np.array(robot_position[:3]))
    return distance < 0.1  # Assuming 0.1m is the threshold for pressing

# Main guidance function
def guidance(state, previous_vars_dict={'maroon_pressed': False, 'green_pressed': False, 'azure_pressed': False}):
    score = 0.0
    vars_dict = previous_vars_dict.copy()
    
    # Get positions of the buttons
    maroon_position = get_position('maroon_button')
    green_position = get_position('green_button')
    azure_position = get_position('azure_button')
    
    # Check the state of the buttons
    if not vars_dict['maroon_pressed']:
        if is_button_pressed(maroon_position, state):
            vars_dict['maroon_pressed'] = True
            score += 1.0  # Increment score for pressing maroon button
    elif not vars_dict['green_pressed']:
        if is_button_pressed(green_position, state):
            vars_dict['green_pressed'] = True
            score += 1.0  # Increment score for pressing green button
    elif not vars_dict['azure_pressed']:
        if is_button_pressed(azure_position, state):
            vars_dict['azure_pressed'] = True
            score += 1.0  # Increment score for pressing azure button

    # Calculate the overall score based on the progress
    score += (vars_dict['maroon_pressed'] + vars_dict['green_pressed'] + vars_dict['azure_pressed']) / 3.0

    return score, vars_dict

\end{lstlisting}

\begin{lstlisting}[language=Python, caption={Guidance code generated for the button-pushing task in the fifth iteration.}, label=lst:fifth_iter]
from grappa.common.perception_functions import get_position, get_size, get_orientation
import numpy as np

# Helper function to check if the button is pressed
def is_button_pressed(button_position, robot_position):
    distance = np.linalg.norm(np.array(button_position) - np.array(robot_position[:3]))
    return distance < 0.05  # Reduced threshold for pressing

# Helper function to calculate movement efficiency
def calculate_movement_score(current_position, target_position):
    distance = np.linalg.norm(np.array(target_position) - np.array(current_position[:3]))
    # Penalize for excessive distance
    if distance > 0.2:  # Arbitrary threshold for excessive distance
        return -0.5  # Strong penalty for being too far
    return max(0, 1 - distance)  # Reward for being close

# Helper function to check orientation using vector mathematics
def is_correct_orientation(button_position, robot_orientation):
    button_vector = np.array(button_position) - np.array([0, 0, 0])  # Assuming button position is in world coordinates
    robot_forward_vector = np.array([np.cos(np.radians(robot_orientation[5])), 
                                      np.sin(np.radians(robot_orientation[5])), 
                                      0])  # Assuming the robot's forward direction is in the XY plane
    angle = np.arccos(np.clip(np.dot(button_vector, robot_forward_vector) / 
                               (np.linalg.norm(button_vector) * np.linalg.norm(robot_forward_vector)), -1.0, 1.0))
    return np.degrees(angle) < 15  # Allow 15 degrees of error

# Main guidance function
def guidance(state, previous_vars_dict={'buttons_pressed': {'maroon': False, 'green': False, 'azure': False}}):
    score = 0.0
    vars_dict = previous_vars_dict.copy()
    
    # Get positions and orientations of the buttons
    maroon_position = get_position('maroon_button')
    green_position = get_position('green_button')
    azure_position = get_position('azure_button')
    
    # Get the current robot position and orientation
    current_position = state
    current_orientation = get_orientation('robot_end_effector')  # Assuming this function exists

    # Button states
    buttons = {
        'maroon': maroon_position,
        'green': green_position,
        'azure': azure_position
    }

    # Check the state of the buttons
    for button, position in buttons.items():
        if not vars_dict['buttons_pressed'][button]:
            if is_button_pressed(position, current_position) and is_correct_orientation(position, current_orientation):
                # Here, you would implement a feedback mechanism to confirm the button press
                # For example: if button_press_successful():
                vars_dict['buttons_pressed'][button] = True
                score += 1.0  # Increment score for pressing the button
            else:
                score += calculate_movement_score(current_position, position)  # Penalize for distance

    # Check if all buttons are pressed
    if all(vars_dict['buttons_pressed'].values()):
        score += 1.0  # Bonus for completing the task

    return score, vars_dict

\end{lstlisting}

\begin{lstlisting}[language=Python, caption={Guidance code generated for the task "push the maroon button, then push the green button, then push the navy button", in iteration 2.}]
from grappa.common.perception_functions import get_position, get_size, get_orientation
import numpy as np

# Helper functions
def distance(point1, point2):
    return np.linalg.norm(np.array(point1) - np.array(point2))

def guidance(state, previous_vars_dict={'maroon_pushed': False, 'green_pushed': False, 'navy_pushed': False}):
    score = 0.0
    vars_dict = previous_vars_dict.copy()
    
    # Get positions of the buttons
    maroon_button_pos = get_position('maroon_button')
    green_button_pos = get_position('green_button')
    navy_button_pos = get_position('navy_button')
    
    # Current end-effector position
    end_effector_pos = state[:3]
    
    # Define thresholds
    push_threshold = 0.05  # 5 cm
    
    if not vars_dict['maroon_pushed']:
        # Move towards maroon button
        dist_to_maroon = distance(end_effector_pos, maroon_button_pos)
        score = 1.0 - dist_to_maroon  # Closer to the button, higher the score
        
        if dist_to_maroon < push_threshold:
            vars_dict['maroon_pushed'] = True
            score += 10  # Bonus for pushing the button
    
    elif not vars_dict['green_pushed']:
        # Move towards green button
        dist_to_green = distance(end_effector_pos, green_button_pos)
        score = 2.0 - dist_to_green  # Closer to the button, higher the score
        
        if dist_to_green < push_threshold:
            vars_dict['green_pushed'] = True
            score += 10  # Bonus for pushing the button
    
    elif not vars_dict['navy_pushed']:
        # Move towards navy button
        dist_to_navy = distance(end_effector_pos, navy_button_pos)
        score = 3.0 - dist_to_navy  # Closer to the button, higher the score
        
        if dist_to_navy < push_threshold:
            vars_dict['navy_pushed'] = True
            score += 10  # Bonus for pushing the button
    
    return score, vars_dict
\end{lstlisting}

\begin{figure}[h]
    \centering
    task: \textbf{``press the maroon button, then press the green button, then press the navy button"}\par\medskip
    \captionsetup{justification=centering}
    \begin{subfigure}[b]{1.0\textwidth}
        \centering
        \includegraphics[width=\textwidth]{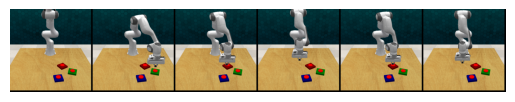}
        \caption{ Act3d with no guidance: the policy fails to press the last button (blue), but manages to correctly approach the first 2 buttons reaching them from above with the gripper closed.}
        \label{fig:no_guidance}
    \end{subfigure}
    \hfill
    \begin{subfigure}[b]{1.0\textwidth}
        \centering
        \includegraphics[width=\textwidth]{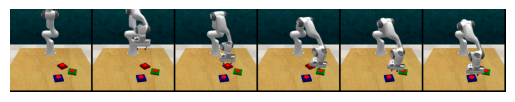}
        \caption{ Guidance only (overwriting the base policy): The sequence of movements is correct, but the initial guidance code doesn't account that the buttons should be approached from above.}
        \label{fig:guidance_only}
    \end{subfigure}
    \hfill
    \begin{subfigure}[b]{1.0\textwidth}
        \centering
        \includegraphics[width=\textwidth]{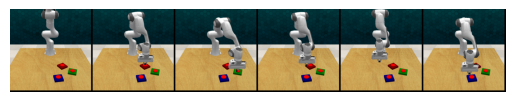}
        \caption{Act3d with 1\% guidance: The modified policy captures both the low-level motion of the pre-trained policy and the high-level guidance provided, successfully pressing the sequence of buttons.}
        \label{fig:0.5_guidance}
    \end{subfigure}
    
    \caption{Sample rollouts of the guidance correcting a failing task.
    }
    \label{fig:rollouts}
\end{figure}



\newpage
\newpage
\clearpage
\begin{lstlisting}[language=Python, caption={Guidance code generated for the task "push the maroon button, then push the green button, then push the navy button", in iteration 2.}]
from grappa.common.perception_functions import get_position, get_size, get_orientation
import numpy as np

# Constants for thresholds
DISTANCE_THRESHOLD = 0.02
GRIPPER_THRESHOLD = 0.01
HEIGHT_OFFSET = 0.05
ORIENTATION_THRESHOLD = 10  # Example threshold for orientation

# Helper function to calculate distance
def calculate_distance(pos1, pos2):
    return np.linalg.norm(np.array(pos1) - np.array(pos2))

# Guidance function
def guidance(state, previous_vars_dict={'button_pressed': False}):
    """
    Guides the robot to press the blue button.
    
    Parameters:
    - state: list of the robot's end-effector position, orientation, and gripper state.
    - previous_vars_dict: dictionary storing the status of conditions relevant to task completion.
    
    Returns:
    - score: float representing how close the robot is to completing the task.
    - button_state: dictionary with the updated status of the button press condition.
    """
    # Get the position of the blue button
    button_position = get_position('blue_button')
        
    # Define the target position slightly above the button
    target_position = [button_position[0], button_position[1], button_position[2] + HEIGHT_OFFSET]
    
    # Calculate the distance to the target position
    distance_to_target = calculate_distance(state[:3], target_position)
    
    # Check if the button is pressed
    button_pressed = distance_to_target < DISTANCE_THRESHOLD and state[6] < GRIPPER_THRESHOLD
    
    # Update button state
    button_state = {'button_pressed': button_pressed}
    
    # Calculate score based on proximity to the target and orientation
    orientation_score = 1.0 if abs(state[5]) < ORIENTATION_THRESHOLD else 0.5
    score = (1.0 / (1.0 + distance_to_target)) * orientation_score
    
    return score, button_state
    
\end{lstlisting}


\begin{lstlisting}[language=Python]
...
def guidance_code(state, hidden_state={"button_pressed": False}):
    #possible future pose of the end-effector
    x,y,z = state[:3]
    robot_pose = [x,y,z]
    
    # get the current pose of the 'red button'
    bt_x, bt_y, bt_z = get_pose('red button')
    
    target_position = [bt_x, bt_y, bt_z + HEIGHT_OFFSET]
    # Calculate the distance to the target position
    dist = calculate_distance(robot_pose, target_position)

    if dist < 0.01:
        new_hidden_state['button_pressed'] = True
    score = dist
    
    return score, new_hidden_state
\end{lstlisting}